\documentclass{article}

\PassOptionsToPackage{numbers,sort&compress}{natbib}
\usepackage[preprint]{neurips_template/neurips_2023}

\usepackage[utf8]{inputenc} %
\usepackage[T1]{fontenc}    %
\usepackage{hyperref}       %
\usepackage{url}            %
\usepackage{booktabs}       %
\usepackage{amsfonts}       %
\usepackage{nicefrac}       %
\usepackage{microtype}      %
\usepackage{xcolor}         %

\usepackage{multirow}
\usepackage{xcolor}
\usepackage{fancyhdr,graphicx}
\usepackage{amssymb,amsthm,amsmath}
\usepackage{booktabs,array} %
\usepackage{thmtools}
\usepackage{thm-restate}
\usepackage[title]{appendix}
\usepackage{tikz}
\usepackage{resizegather}
\usepackage{wrapfig}
\usepackage{enumitem,kantlipsum}
\usepackage{mathtools}
\usepackage{nccmath}

\newtheoremstyle{mystyle}
{6pt} %
{\topsep} %
{} %
{} %
{\bfseries} %
{.} %
{.5em} %
{} %

\theoremstyle{definition}
\newtheorem{definition}{Definition}
\newtheorem{assumption}{Assumption}
\theoremstyle{plain}

\newtheorem{theorem}{Theorem}

 \usepackage{xspace}

\usepackage{dsfont}
\usepackage{stmaryrd}

\renewcommand{\phi}{\varphi}

\newcommand{\rank}{\operatorname{rank}}

\newcommand{\norm}[1]{\Vert {#1} \Vert}

\newcommand{\loss}{\mathcal{L}}
\newcommand{\funcF}{\mathcal{F}}

\newcommand{\algoname}{InRank\xspace}

\usepackage{amsmath,amsfonts,bm}

\def\eqref#1{equation~\ref{#1}}

\def\1{\bm{1}}

\DeclareMathAlphabet{\mathsfit}{\encodingdefault}{\sfdefault}{m}{sl}
\SetMathAlphabet{\mathsfit}{bold}{\encodingdefault}{\sfdefault}{bx}{n}

\def\sR{{\mathbb{R}}}

\newcommand{\R}{\mathbb{R}}

\usepackage{graphicx}
\usepackage{pifont}
\usepackage{amssymb}
\usepackage{multicol}
\usepackage{tablefootnote}
\usepackage[para,online,flushleft]{threeparttable}

\usepackage{setspace}

\usepackage{algorithm}
\usepackage{algpseudocode}

\usepackage[many]{tcolorbox}

\usepackage{subcaption}

\title{InRank: Incremental Low-Rank Learning}

\author{%
Jiawei Zhao\thanks{Equal contribution, names in alphabetical order.}\\
California Institute of Technology\\
\small{\texttt{jiawei@caltech.edu}} \\
\And
Yifei Zhang\footnotemark[1]\\
University of Wisconsin-Madison\\
\small{\texttt{yzhang2536@wisc.edu}} \\
\And
\hspace{0.3in} Beidi Chen\\
\hspace{0.2in}Carnegie Mellon University\\
\hspace{0.2in}\small{\texttt{beidic@andrew.cmu.edu}} \\
\And
\hspace{0.4in} Florian Sch{\"a}fer\\
\hspace{0.3in} Georgia Institute of Technology\\
\hspace{0.3in} \small{\texttt{florian.schaefer@cc.gatech.edu}} \\
\And
Anima Anandkumar \\
California Institute of Technology\\
NVIDIA\\
\small{\texttt{anima@caltech.edu}} \\
}

\begin{document}

\maketitle
\begin{abstract}

   The theory of greedy low-rank learning (GLRL) aims to explain the impressive generalization capabilities of deep learning.
   It proves that stochastic gradient-based training implicitly regularizes neural networks towards low-rank solutions through a gradual increase of the rank during training.
   However, there is a gap between theory and practice since GLRL requires an infinitesimal initialization of the weights, which is not practical due to the fact that it is a saddle point.
   In this work, we remove the assumption of infinitesimal initialization by focusing on \emph{cumulative weight updates}. We prove the cumulative weight updates follow an incremental low-rank trajectory for arbitrary orthogonal initialization of weights in a three-layer linear network. 
   Empirically, we demonstrate that our theory holds on a broad range of neural networks (e.g., transformers) and standard training algorithms (e.g., SGD, Adam).  However, existing training algorithms do not exploit the low-rank property to improve computational efficiency as the networks are not parameterized in low-rank.  To remedy this, we design a new training algorithm \emph{Incremental Low-Rank Learning} (\algoname), which explicitly expresses  cumulative weight updates  as  low-rank matrices while incrementally augmenting their ranks during training.
   We evaluate \algoname on GPT-2, and our results indicate that \algoname achieves comparable prediction performance as the full-rank counterpart while requiring at most 33\% of the total ranks throughout training.
   We also propose an efficient version of \algoname that achieves a reduction of 37\% in total training time and 36\% in model size when training GPT-medium on WikiText-103 from scratch. Repository publicly available on Github: \href{https://github.com/jiaweizzhao/InRank}{https://github.com/jiaweizzhao/InRank}.

\end{abstract}
\section{Introduction}

The generalization ability of deep neural networks continues to intrigue researchers since the classical theory is not applicable in the over-parameterized regime, where there are more learnable parameters than training samples.
Instead, efforts to understand this puzzle are based on the belief that first-order learning algorithms (e.g., stochastic gradient descent) implicitly bias the neural networks toward simple solutions. 

For instance, it has been shown that stochastic gradient descent implicitly minimizes the rank of solutions during training \citep{arora_implicit_2019}.
Recent theoretical studies have further demonstrated one of its training characterizations - Greedy Low-Rank Learning (GLRL) \citep{li_towards_2021,jacot_saddle--saddle_2022}.
GLRL characterizes the trajectory of stochastic gradient descent, which performs a rank-constrained optimization and greedily increases the rank whenever it fails to reach a global minimizer.  

However, one major drawback is that the GLRL theory requires the assumption of 
infinitesimal initialization, which is impractical as gradient descent cannot effortlessly escape from the saddle point at zero, unless the noise is large enough.
Therefore, a generalized notion of GLRL under practical initialization is needed to bridge the gap between theory and practice.

{\bf In this work}, we generalize the theory of GLRL by removing the requirement of infinitesimal initialization. To do this, we focus on characterizing the trajectories of a new set of quantities, \emph{cumulative weight updates}, instead of weight matrices. Cumulative weight updates do not include the initialization values, and only incorporate the rest of the updates of the weight matrices during training. This allows us to remove the requirement of infinitesimal initialization in GLRL.

We establish incremental rank augmentation of cumulative weight updates during training under arbitrary orthogonal initialization of the weights. 
This new formulation proves that low-rank learning can be extended to non-zero initialization, where the singular vector with a larger value in the associated target matrix is learned exponentially faster.
We prove this relationship by following the work of \citet{saxe_exact_2014} to analyze the evolution of each mode (singular vector) independently, which can be achieved by ensuring orthogonality over the weights matrices and inputs in a three-layer linear network.

Empirically, we further demonstrate that standard networks (e.g., transformers) and training algorithms (e.g., SGD, Adam) follow low-rank learning trajectories on the cumulative weight updates, under standard weight initialization.
However, current algorithms can not exploit the low-rank property to improve computational efficiency as the networks are not parameterized in low-rank.

\begin{figure}[t!]
    \centering
    \includegraphics[width=1.0\linewidth]{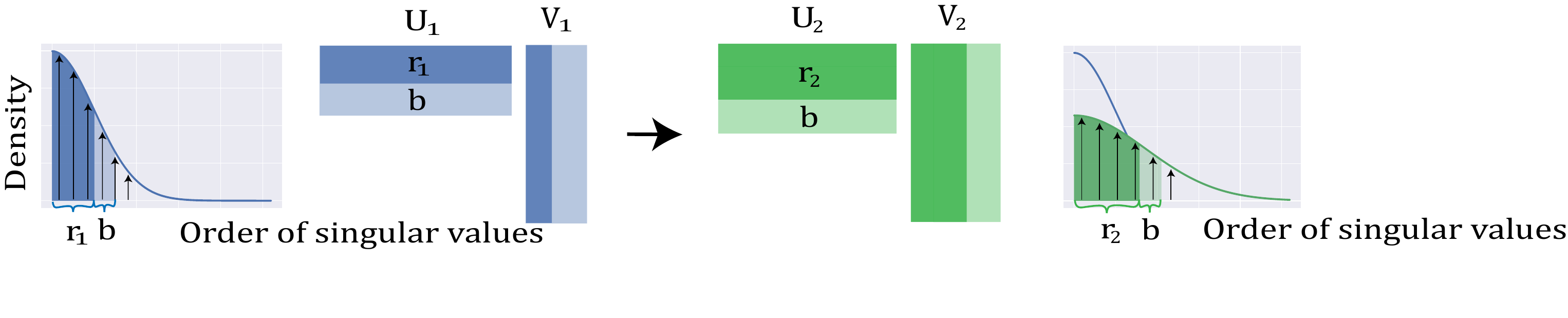}
    \caption{\textbf{Incremental Low-Rank Learning from iteration $t_1$ to $t_2$.} $U$ and $V$ represent any factorized layer. Density plots indicate the strength of each singular vector (normalized by the total strengths). Solid areas represent how much information in the spectrum is explained by the current rank $r_t$ at iteration $t$. From iteration $t_1$ to $t_2$, \algoname adds $r_{2} - r_{1}$ additional ranks to ensure the ratio of the explained information is greater than a certain threshold $\alpha$.}
    \label{fig:algo-framework}
    \vskip -0.1in
\end{figure}

To address this, we propose \emph{Incremental Low-Rank Learning} (\algoname), which parameterizes the cumulative weight updates in low-rank while incrementally augmenting its rank during training, as illustrated in Figure~\ref{fig:algo-framework}.
\algoname adds a new batch of modes whenever a certain quantity, known as the explained ratio,
exceeds a certain threshold. 
The explained ratio represents the amount of information in the underlying spectrum that the current rank can explain.
A low explained ratio indicates that the current rank is inadequate to represent the spectrum, necessitating the addition of more modes.

\algoname is capable of identifying \emph{intrinsic rank} of networks during training.
The intrinsic rank of a neural network is defined as the minimum sufficient rank that trains the network from scratch without sacrificing performance.
The capability of finding the intrinsic rank addresses the challenge of pre-defining the fixed ranks in training low-rank neural networks, which requires expensive hyperparameter tuning.
An inappropriate selection of rank may either limit model capacity, hinder the learning process, or result in excessive computation, thereby negating the benefits of low-rank factorization. We further improve computational efficiency by applying \algoname only in the initial phase of training. 
This approach mitigates the computational cost induced by expensive SVD operations in \algoname, while maintaining comparable accuracy as the full-rank models.

\textbf{Our contributions are summarized as follows:}
\begin{enumerate}[leftmargin=*]
    \item We generalize the theory of GLRL to arbitrary orthogonal initialization by establishing incremental rank augmentation of cumulative weight updates. Empirically, we demonstrate that our theory holds on a broad
    range of neural networks and standard training algorithms. 
    \item We propose \algoname, which can find the intrinsic rank of networks during training. To our best knowledge, \algoname is the first practical algorithm that achieves incremental low-rank learning in large-scale training of neural networks.
    \item \algoname maintains prediction performance equivalent to full-rank counterparts but requires a maximum of 33\% total ranks when evaluating \algoname on WikiText-103 using GPT-2 models.
    \item We further enhance the training efficiency of \algoname. This efficient variant decreases the training time by 37\% and reduces model size by 36\% with a 10\% reduction of memory usage when training GPT-medium from scratch.
\end{enumerate}

\section{Related Work}

Implicit regularization has been well studied to explain excellent generalization in neural networks \citep{gunasekar_characterizing_2018,rahaman_spectral_2019}.
Implicit rank regularization stands out among the diverse aspects of implicit regularization, which demonstrates that a neural network minimizes its rank implicitly during training \citep{arora_implicit_2019,gissin_implicit_2019}.
Further research has corroborated that neural networks pursue a greedy low-rank learning strategy under infinitesimal initialization \citep{razin_implicit_2021,jacot_saddle--saddle_2022,li_towards_2021}.
However, the practical advantages of such an approach remain unexplored, predominantly due to the challenges of deviating from the infinitesimal initialization assumption.

Low-rank training and other structured pruning methods aim to promote structured sparsity within neural networks (NNs) throughout the training process, enabling substantial computational acceleration \citep{you_drawing_2022,dao_monarch_2022}.
The low-rank training technique has proven effective for training low-rank neural networks from scratch \citep{ioannou_training_2016,yang_learning_2020,schotthofer_low-rank_2022}.
Nonetheless, this method often necessitates extra hyperparameters, such as the rank of the factorization, which can be challenging to determine accurately, and thus it requires careful tuning.

\citet{idelbayev_low-rank_2020} propose the LC compression method that explicitly integrates the learning of low-rank factors into the training process, despite its computational intensity. 
More recently, \citet{wang_cuttlefish_2023} introduce Cuttlefish, a low-rank training method capable of automatically determining the factorization rank in the early stages of training. 
However, Cuttlefish requires a pre-set full-rank initialization and lacks a theoretical comprehension of its low-rank behavior, unlike our proposed \algoname.

Moreover, low-rank training has been employed for fine-tuning large-scale pre-trained models \citep{hu_lora_2021}, and for reducing communication overhead in distributed training \citep{vogels_powersgd_2019,wang_pufferfish_2021_new}. 
\citet{liMeasuringIntrinsicDimension2018} adopt the low-rankness in cumulative weight updates to measure the intrinsic dimension of objective landscapes.
The concept of incremental learning has been examined within the context of learning partial differential equations using neural networks, such as parameter expansion in the frequency domain \citep{zhao_incremental_2022}, and increasing the complexity of the underlying PDE problem \citep{huang_pinnup_2021}.

\begin{algorithm}[h!]
    \caption{Greedy Low-Rank Learning (GLRL)}
    \label{alg:glrl}
    \begin{algorithmic}[1]
        \Require $C$ is a convex cost, $A_{\theta}$ is the product matrix: $A_{\theta} = W^1 ... W^L$, and let $\epsilon, \eta, T >0$
        \State Compute the first singular vector of $\nabla C(0)$: $u, s, v \gets \mathrm{SVD}_1(\nabla C(0))$
        \State Initialize parameters and network width: $\theta \gets (-\epsilon v^T, \epsilon, \ldots, \epsilon u)$, $w \gets 1$
        \While{$C(A_{\theta}) < C_{min} + \epsilon$} %
        \State Train width-$w$ deep linear network for $T$ steps using SGD with learning rate $\eta$
        \State Compute the first singular vectors of $\nabla C(A_{\theta})$: $u, s, v \gets \mathrm{SVD}_1(\nabla C(A_{\theta}))$
        \State Expand network width: $w \gets w + 1$
        \State Initialize additional parameters:
        $$ \theta \leftarrow\left(\left(\begin{array}{c}
            W^1 \\
            -\epsilon v^T
        \end{array}\right),\left(\begin{array}{cc}
            W^2 & 0 \\
            0 & \epsilon
        \end{array}\right), \ldots,\left(\begin{array}{cc}
            W^L & \epsilon u
        \end{array}\right)\right) 
        $$
        \EndWhile
    \end{algorithmic}
\end{algorithm}

\section{Preliminary: Greedy Low-Rank Learning}

In this section, we first introduce greedy low-rank learning (GLRL) and its practical limitations. 
We wish to train the function $\funcF(x)$ to learn a particular input-output map given a set of $P$ training samples $(x_{\mu},y_{\mu}) \in \sR^{N_{x} \times N_{y}}$, where $\mu = 1,...,P$. 
Training is accomplished by minimizing the squared error $\loss = \frac{1}{2} \sum_{\mu=1}^{P} \norm{y_{\mu} - \funcF(x_{\mu})}_2^2$ using gradient descent with a step size $\eta$.

We first model $\funcF(x)$ to be a deep linear network: $\funcF(x) =  W^{L} \, ... \, W^{1} \, x$, where $W^{l} \in \sR^{N_{h} \times N_{h}}$ for $l \in 1,...,L$. We let $A_{\theta} = W^{L} \, ... \, W^{1}$ denote the product matrix of the network, and $\theta$ denote the whole parameter vector.
Thus, we also denote the training error as $C(A_{\theta})$ where $C$ is a convex error (e.g., the squared error).

The following theorem characterizes the implicit rank regularization behavior of gradient descent under infinitesimal initialization.
\begin{theorem}[Greedy Low-Rank Learning, informal]
    \label{thm:glrl}
    If initialize $W^1,...,W^L$ to be infinitesimal, then the product matrix $A_{\theta}$ follows a greedy low-rank learning trajectory, such that the gradient descent first searches over a rank-1 subspace of $A_{\theta}$, and then greedily increases the rank by one whenever it fails to reach a local minimizer.
\end{theorem}
Theorem \ref{thm:glrl} characterizes the trajectory of gradient descent, which performs a rank-constrained optimization and greedily relaxes the rank restriction until it finds a local minimizer.

Inspired by this implicit low-rank trajectory, the greedy low-rank learning (GLRL) algorithm is proposed to capture this implicit behavior explicitly \citep{li_towards_2021}. As shown in Algorithm~\ref{alg:glrl}, GLRL incrementally increases the rank of the weight matrices in a deep linear network and initializes the additional rows and columns based on the top singular vector of the current matrix derivative.

Although the GLRL algorithm provides a theoretical understanding of implicit rank regularization, it has some practical drawbacks. 
One notable limitation is the infinitesimally small initialization, which leads to slow convergence and makes it difficult to apply the algorithm in large-scale settings. 
In addition, GLRL is only applicable to linear networks as it highly relies on the product matrix $A_{\theta}$.
This makes it inapplicable to practical neural networks with non-linear activation functions.

\section{Cumulative Weight Updates follow Low-Rank Learning Trajectory}

In order to generalize GLRL beyond infinitesimal initialization, we focus on \textit{cumulative weight updates} that characterize GLRL for any regular initializations. We define the cumulative weight updates as follows:

\begin{definition}[Cumulative Weight Updates]
    The \emph{cumulative weight updates} $d_{t}$ at iteration $t$ is defined as the difference between the current parameterization $w_{t}$ and initialization $w_{0}$ in the parameter space, such that
    \begin{equation}
        d_{t} = w_{t} - w_{0} = \sum_{i=1}^{t} \Delta w_{i}.
    \end{equation}    
\end{definition}

The cumulative weight updates $d_{t}$ have been widely studied in the literature, especially in the field of distributed training \citep{vogels_powersgd_2019}, as it is known to exhibit low-rank properties.  

This is attributed to the fact that $d_{t}$ is a summation of updates to the weights $\Delta w_{i}$, with each update being determined by the learning algorithm and current gradient $g_t$. 
Gradient $g_t$ has been shown to possess low-rank properties, which has been exploited to reduce communication costs in distributed training through low-rank compression \citep{vogels_powersgd_2019,wang_pufferfish_2021_new}.

We theoretically prove that the cumulative weight updates $d_{t}$ follow a low-rank learning trajectory, even when the initialization is not infinitesimal. We continue to focus on a linear network and analyze the difference of the product matrix $D_t = A_t - A_0$ (which can be viewed as the cumulative weight updates of the product matrix). Our goal is to demonstrate that $D_t$ exhibits an exponential rank increase even when the initial weights are not close to zero. Our analysis builds upon the work of \citet{saxe_exact_2014}, which studies training dynamics under orthogonal inputs.

\begin{assumption}[Orthogonal Inputs]
    We assume the inputs are orthogonal, i.e., $x_i^T x_j = 0$ for $i \neq j$.
\end{assumption}

Consider the input-output correlation matrix:
\begin{equation}
    \Sigma^{yx} = \sum_{\mu=1}^P y_{\mu} x_{\mu}^T = U^{yy} S^{yx} V^{xx}=\sum_{\alpha=1}^{N_x} s_\alpha u_\alpha v_\alpha^T,
\end{equation}
where $U^{yy}$ and $V^{xx}$ represent the left and right singular vectors of $\Sigma^{yx}$, and $S^{yx}$ denotes its singular value matrix. The singular values are ordered such that $s_1 \geq s_2 \geq \dots \geq s_{N_x}$.

We analyze a 3-layer linear network where $y = W^2 W^1 x$, $W^1 \in \R^{N_h \times N_x}$ and $W^2 \in \R^{N_y \times N_h}$ are the weight matrices of the first and second layers, respectively, and $N_h < N_x, N_y$. After training, the converged network should satisfy:
\begin{equation}
    W^2 W^1 =\sum_{\alpha=1}^{N_h} s_\alpha u_\alpha v_\alpha^T,
\end{equation}
which is the closest rank-$N_h$ approximation to $\Sigma^{yx}$. 
To further analyze its trajectory, we assume that the weights are initialized as $W^{2}_0=U^{yy} M^2 O^T, W^{1}_0=O M^1 V^{xx^T}$, where $M^2, M^1$ are diagonal matrices, and $O$ is an arbitrary orthogonal matrix. We have the following theorem for the training evolution of $D_t$:

\begin{theorem}
    \label{thm:network-difference}
    For any orthogonal matrix $O$ and scaled diagonal matrices $M^2$ and $M^1$, each singular value $u_f(t)$ in $D_t$ at iteration $t$ follows the trajectory:
    \begin{equation}
        u_f(t)=\frac{s e^{2 s t / \tau}}{e^{2 s t / \tau}-1+s / u_0} - u_0,
    \end{equation}
    where $s$ is the target singular value in $\Sigma^{yx}$, $u_0$ is the initial value determined by $M^2$ and $M^1$, and $\tau$ is a constant.
\end{theorem}
$W^{2}_0 W^{1}_0$ ensures that each mode $\alpha$ is learned independently right from the beginning of training, enabling us to analyze the learning trajectory of each mode separately.
The diagonal matrices $M^2$ and $M^1$ control the scale of the initial weights, i.e., the initial value $u_0$ of each mode $\alpha$.
Consequently, a larger $u_0$ that is closer to $s$ accelerates the learning speed.
We provide comprehensive proof of the theorem in the appendix for further clarity.

The sigmoid function in Theorem~\ref{thm:network-difference} exhibits a sharp transition from a state of no learning to full learning, with the transition point determined by the initial value $u_0$ and $s$.
This indicates that if the target singular values $s$ are distinct enough (given $s >> u_{0}$), each $u_f(t)$ will follow an independent sigmoid trajectory, permitting ranks to be learned sequentially and independently.

\begin{figure}[t!]
    \centering
    \includegraphics[width=1.0\linewidth]{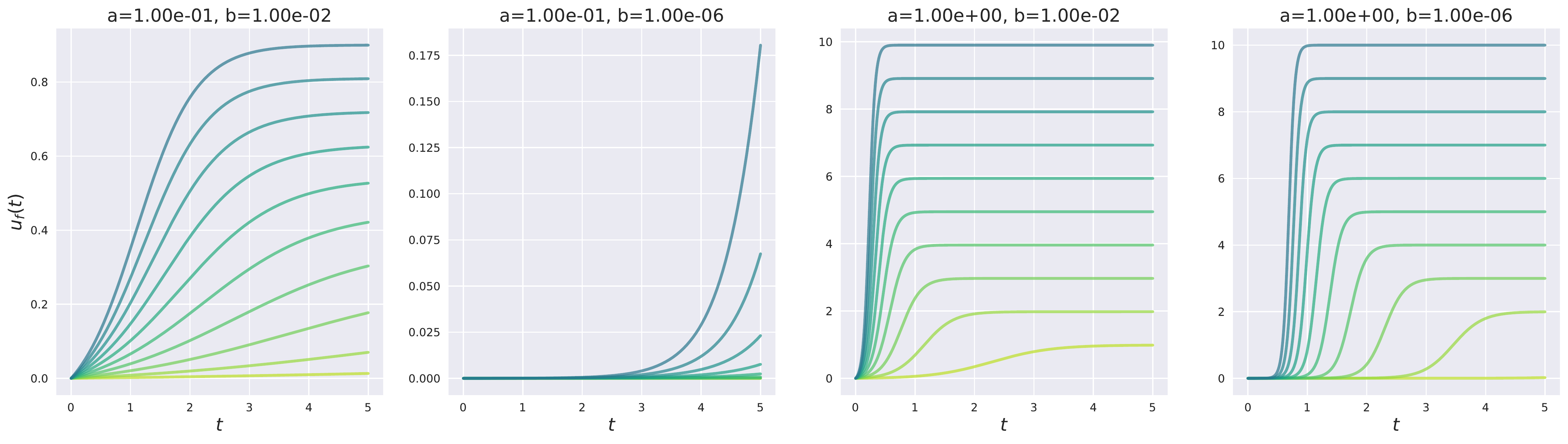}
    \caption{$u_f(t)$ follows low-rank learning trajectory regardless of $s$ and $u_0$. We generate a set of $s$ given $s_i = a \times i, i=1,...,10$ while varying $a$ from $0.1$ to $1.0$. We also generate a set of $u_0$ given $u_0 \sim \mathcal{N}(0, b^2)$. Darker colors indicate singular vectors with higher strengths.}
    \label{fig:theory_verification}
\end{figure}

To validate this, we carry out an empirical simulation using different sets of $u_0$ and $s$. 
As illustrated in Figure~\ref{fig:theory_verification}, under various scales of initialization, the evolution of $u_f(t)$ consistently adheres to the low-rank learning trajectory.
We note that analyzing weights $W^2 W^1$ directly under infinitesimal initialization in \citet{li_towards_2021} can be viewed as a special case of analyzing $D_t$ here.

\begin{figure}[h!]
    \centering
    \includegraphics[width=1.0\linewidth]{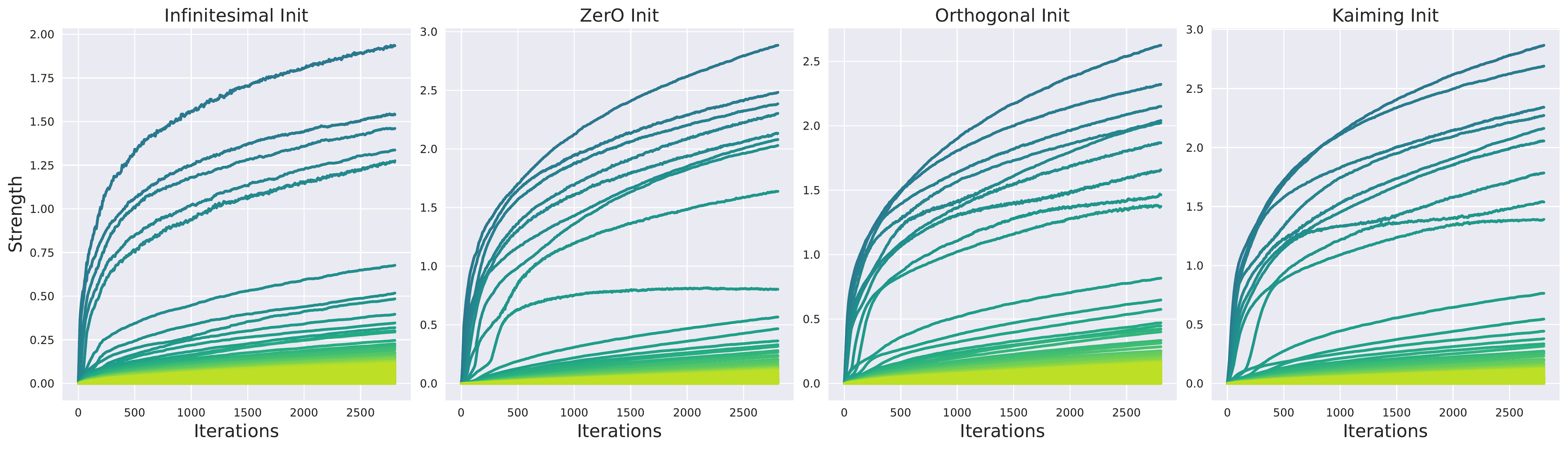}
    \caption{The evolutions of all singular vectors of cumulative weight updates $D_t$ over training under different initializations. They are evaluated on the training of a 2-layer perceptron on Fashion MNIST. Darker colors indicate singular vectors with higher strengths.}
    \label{fig:spectrum_over_training}
  \end{figure}

Shifting our focus to practical non-linear networks, we analyze the difference of layer-wise weight matrix $D_t = W^{l}_{t} - W^{l}_{0}$ for $l=1,...,L$ instead of the product matrix $D_t = W^{L}_{t}...W^{1}_{t} - W^{L}_{0}...W^{1}_{0}$.
We also extend our evaluation to more practical cases with modern weight initialization methods. 
As shown in Figure~\ref{fig:spectrum_over_training}, cumulative weight updates $D_t$ follow the greedy low-rank learning trajectory even under regular initializations, including Orthogonal, ZerO, and Kaiming methods \citep{saxe_exact_2014,<kaiming>,zhao_zero_2021}.

We further verify our theory on a broad range of neural networks (e.g., transformers) and standard training algorithms (e.g., SGD, Adam), as shown in the appendix.
This observation motivates us to design an efficient incremental learning algorithm that leverages the properties of cumulative weight updates.

\section{Incremental Learning}

\begin{algorithm}[h!]
    \caption{Incremental Low-Rank Learning (\algoname)}
    \label{alg:incremental-learning}
    \begin{algorithmic}[1]
        \Require $L(W_t)$ is the cost of total weights $W=(W^1,...,W^L)$ at iteration $t$, $W_t^{l} = W_0^{l} + U_t^{l} V_t^{l}$ for each layer $l$ ($W_0^{l}$ is not trainable) and let $r, b, \alpha, \epsilon, \eta, T > 0$
        \State Initialize $W_0^{l}$ using standard initialization, and set $U_0^{l}, V_0^{l}$ to 0
        \State Compute the top $(2+b)$ singular vectors: $u^{l}, s^{l}, v^{l} \gets \mathrm{SVD}_{(2+b)}(\frac{\partial L(W_0)}{\partial W_0^{l}})$
        \State Initialize factorized weights with small $\epsilon$ : $U_0^{l} \gets - \epsilon v^{l}$, $V_0^{l} \gets \epsilon u^{l}$, and $r^{l}_{0} \gets 2$
        \For{$t=1,2,3,...,T$}
        \State Train low-rank network and update $U_t^{l} V_t^{l}$ using SGD with learning rate $\eta$
        \State Compute the top $(r^{l}+b)$ singular vectors: $u^{l}, s^{l}, v^{l} \gets \mathrm{SVD}_{(r^{l}+b)}(U_t^{l} V_t^{l})$
        \State Increment $r^{l}_{t}$ to $r^{l}_{t+1}$ until the explained ratio $g(U_t^{l} V_t^{l},r^{l}_{t+1},b) \geq \alpha$
        \State Initialize additional parameters: $U_{t+1}^{l} \gets [U_t^{l}, U^{*}]$, $V_{t+1}^{l} \gets [V_t^{l}, V^{*}]$, where $U^{*} \in \sR^{p^{l} \times (r^{l}_{t+1} - r^{l}_{t})}$ and $V^{*} \in \sR^{(r^{l}_{t+1} - r^{l}_{t}) \times q^{l}}$ are randomly initialized with small values
        \EndFor
    \end{algorithmic}
\end{algorithm}

Motivated by the previous findings, we propose an incremental low-rank learning algorithm that leverages the implicit low-rank learning trajectory in practice.
To explicitly represent the cumulative weight updates, we parametrize the weight matrix $W^{l}$ at any layer $l$ as follows:
\begin{equation}
    W^{l} = W_0^{l} + D^{l},
\end{equation}
where $W_0^{l}$ is the initial matrix and $D^{l}$ is the summation of weight updates. 
Since $D^{l}$ exhibits low-rank properties, we can factorize it as $D^{l} = U^{l} V^{l}$, resulting in:
\begin{equation}
    \label{eq:weight-param}
    W^{l} = W_0^{l} + U^{l} V^{l}, 
\end{equation}
where $U^{l} \in \mathbb{R}^{p^{l} \times r^{l}}$ and $V^{l} \in \mathbb{R}^{r^{l} \times q^{l}}$ are the factorized matrices, and $r^{l}$ is the rank of $D^{l}$.

To emulate the implicit low-rank learning, we train factorized matrices $U^{l} V^{l}$ with an initially small rank $r^{l}$, subsequently increasing the rank (i.e., the matrix size) throughout the training process.

A crucial challenge lies in determining how to increase the rank $r^{l}$ during training. 
An inappropriate choice of rank may either lead to insufficient model capacity, hinder the learning process, or result in excessive computation, negating the benefits of low-rank factorization.

To address this, we propose a novel method for dynamically identifying when a rank increase is necessary, based on measuring the representability of the current rank $r^{l}$. Inspired by \citet{zhao_incremental_2022}, we define \emph{explained ratio} by explained variation:
\begin{equation}
    \label{eq:explained-ratio}
    g(M,r^{l},b) = 1-\frac{var([s_{r+1}^{l}, ..., s_{r+b}^{l}])}{var([s_{1}^{l}, ..., s_{r+b}^{l}])},
\end{equation}
where $s_{i}^{l}$ is the $i$-th singular value of a matrix $M$, and $b$ is a buffer size used to encompass a broader spectrum for determination; $var$ operator computes a list of singular values.
The explained ratio $g$ quantifies the representability of the current rank $r^{l}$ in the truncated spectrum (of size $(r^{l}+b)$) of $M$.
A low explained ratio $g$ indicates that the existing rank $r^{l}$ cannot sufficiently represent the truncated spectrum, necessitating an increase in $r^{l}$ to incorporate more useful modes. 

We let $M = U^{l} V^{l}$ for each layer $l$ and compute the explained ratio $g(U^{l} V^{l},r^{l},b)$ at each iteration (can be relaxed each $k$ iterations in practice).
By predefining an appropriate threshold $\alpha$ and ensuring that $g(U^{l} V^{l},r^{l},b)$ remains larger than $\alpha$ during training, the rank $r^{l}$ can automatically increase when needed. 
This process is illustrated in Figure~\ref{fig:algo-framework}.

It is worth noting that $b$ buffer ranks serve to provide a wider spectrum range, but their corresponding singular vectors may be less useful. These buffer ranks can be discarded by fine-tuning in the post-training stage. The full algorithm is detailed in Algorithm~\ref{alg:incremental-learning}.

\section{Evaluation}
In this section, we conduct a comprehensive evaluation of our proposed \algoname algorithm on GPT-2.

Our method particularly focuses on the fully-connected layers in the models, where we substitute the conventional weight parameterization with our relative parameterization as described in Equation~\ref{eq:weight-param}. This operation involves applying \algoname to the resulting low-rank factorized matrices. Notably, our approach is not exclusively limited to fully-connected layers. It bears the flexibility to be extended to various types of layers, including convolution and self-attention layers. However, to maintain the focus on our current research, we leave this promising exploration for future work.

We benchmark the effectiveness of our method mainly on Generative Pre-trained Transformer 2 (GPT-2), a model widely used in language tasks.
In our experiment, we apply \algoname to all the MLP layers in GPT-2 and assess the training of GPT-2 from scratch on the WikiText-103 dataset.

We fix the hyperparameters of \algoname across all experiments and different models, including an initial rank of $r_0 = 2$, a buffer size of $b=100$, and a threshold of $\alpha=0.9$.
We find both values $r_0$ and $b$ are insensitive to the performance of \algoname, and we will discuss the selection of the threshold $\alpha$ in the following section.

\subsection{Automatic Rank Determination}

\begin{figure}[h!]
    \centering
    \includegraphics[height=0.4\textwidth, width=0.8\textwidth]{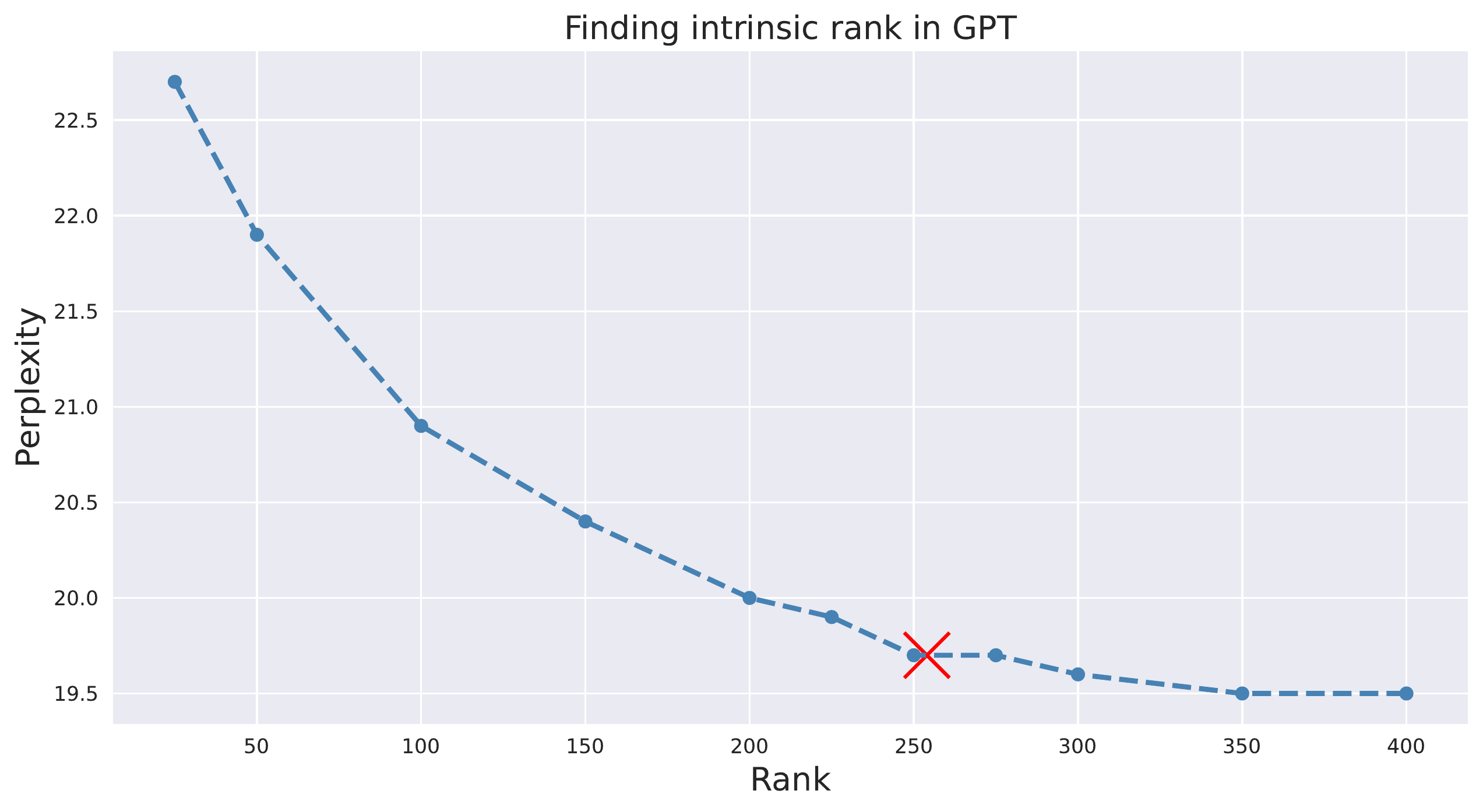}
    \caption{Identifying intrinsic rank in GPT-small on WikiText-103. The cross marker signifies the rank determined by \algoname. The rank varies from 25 to 400. } 
    \label{fig:varying_rank}
\end{figure}

A key finding from our evaluation is that \algoname can automatically find the intrinsic rank of the model during training, facilitated by the automatic rank determination feature in cumulative weight updates.
Figure~\ref{fig:varying_rank} demonstrates that the rank identified by \algoname aligns with the intrinsic rank discovered by costly sweeping across a wide range of ranks.
This capability could potentially eliminate the need for the laborious and time-intensive process of tuning the rank hyperparameters for training low-rank networks.

\subsection{\algoname-Efficient}

We aim to improve the efficiency of \algoname.
We find the rank increment mostly occurs during the early stages of training, remaining relatively stable thereafter. 
This observation suggests that the initial training phase can sufficiently infer the intrinsic rank of the model, corroborating the findings of previous work~\cite{wang_cuttlefish_2023}.

This motivates us to apply \algoname only in the early stage to determine an appropriate rank for low-rank training, fixing its rank afterward.
We denote this variant as \algoname-Efficient.
The conventional \algoname is computationally expensive due to the $O(n^3)$ cost of the SVD operation for a matrix of size $n \times n$. 
On the other hand, \algoname-Efficient reduces the computational burden by only applying \algoname during the initial training stage.
In the remaining evaluation on GPT-2, \algoname-Efficient is only applied for the first epoch.

In the \algoname-Efficient approach, once we determine the optimal rank $r^{*}$ for $UV$ using \algoname, we reparameterize the weights as a rank-$r^{*}$ factorization of $U^{*}V^{*}$ only, such that $W = U^{*}V^{*}$ where $U^{*}V^{*} = W_0 + UV$.
This reduces additional computational cost of representing $W_0$ in the rest of training.

Moreover, we can enhance efficiency by discarding the buffer size $b$ by once SVD operation when the optimal rank has been determined.
We provide further details of \algoname-Efficient in the appendix.

\subsection{Comparison}

We compare \algoname-Efficient with a full-rank baseline using different sizes of GPT models on the WikiText-103 dataset.
For each type of model, the \algoname-Efficient and full-rank baseline are trained with the same hyperparameters, including the learning rate, weight decay, and the number of epochs.
We use the AdamW optimizer to train for 100 epochs while maintaining gradient accumulation to reach an effective batch size of 512.
All experiments are run using the same computational setting with 8 NVIDIA$^{\circledR}$ V100 GPUs 32GB or 8 NVIDIA$^{\circledR}$ A100 GPUs 80GB. We provide further details required to replicate our experiments in the appendix.

As shown in Table~\ref{tab:main_benchmark}, \algoname-Efficient achieve perplexity comparable even superior for GPT-medium and GPT-large to the full-rank baseline while requiring at most 33\% of the total rank.
The rank is calculated as the average rank across all weight matrices in the model. We also measure several efficiency metrics to compare the computational efficiency of different methods.
Specifically, we measure the total training time, memory usage, model size, and number of parameters required for training.

\begin{table}[b!]
    \centering
    \caption{Performance comparison of different methods.}
    \begin{tabular}{llcccccc}
        \toprule
        \textbf{Model} & \textbf{Method} & \textbf{PPL} & \textbf{Rank} & \textbf{Runtime} & \textbf{Memory} & \textbf{Model Size} & \textbf{Params} \\
        \midrule
        GPT-small & Baseline & \textbf{19.3} & 768 & 23.9h & 27GB & 248Mb & 124M \\
                  & \algoname-Efficient & 19.6 & \textbf{254} & \textbf{19.1h} & \textbf{24.7GB} & \textbf{182Mb} & \textbf{91.2M} \\
        GPT-medium & Baseline & 20.6 & 1024 & 28.7h & 73GB & 1.4KMb & 354M \\
                   & \algoname-Efficient & \textbf{20.2} & \textbf{286} & \textbf{18.0h} & \textbf{65.1GB} & \textbf{895Mb} & \textbf{223M}  \\
        GPT-large & Baseline & 20.9 & 1280 & 55.6h & 71GB& 3.1KMb & 774M \\
                  & \algoname-Efficient & \textbf{20.4} & \textbf{312} & \textbf{34.6h} & \textbf{58.2GB} & \textbf{1.8KMb} & \textbf{445M}  \\
        \bottomrule
    \end{tabular}
    \label{tab:main_benchmark}
\end{table}

Notably, \algoname-Efficient significantly reduces both computational cost and memory usage with a reduction in memory usage.
For instance, when compared to the baseline on GPT-medium, \algoname-Efficient reduces the total training time by 37\% and model size by
\begin{minipage}{0.47\textwidth}
36\% with a 10\% reduction in memory usage. Moreover, \algoname-Efficient demonstrates greater efficiency benefits with larger models. In the case of GPT-large, \algoname-Efficient reduces 75\% of the total rank, resulting in a reduction of 38\% in training time and 42\% in model size, along with a 18\% reduction in memory usage.
\setlength{\parskip}{0.50\baselineskip}

Additionally, we fix the rank as the number auto-determined by \algoname-Efficient at the beginning and train GPT-2 from the scratch. 
As shown in Table~\ref{tab:comparison_with_fixed_rank}, InRank-Efficient even achieves lower testing perplexity compared to the fixed rank method, suggesting that incremental low-rank learning serves as an implicit regularization to improve the generalization.
\end{minipage}
\hfill
\begin{minipage}{0.5\textwidth}
    \raggedleft
    \centering
    \captionof{table}{Performance comparison with fixed rank.}
    \begin{tabular}{llc}
        \toprule
        \textbf{Model} & \textbf{Method} & \textbf{PPL}\\
        \midrule
        GPT-small & Fixed Rank & 19.7 \\
        & \algoname-Efficient & \textbf{19.6} \\
        GPT-medium & Fixed Rank & 20.5 \\
        & \algoname-Efficient & \textbf{20.2} \\
        GPT-large & Fixed Rank & 20.8 \\
        & \algoname-Efficient & \textbf{20.4} \\
        \bottomrule
    \end{tabular}
    \label{tab:comparison_with_fixed_rank}
\end{minipage}

\subsection{Selection of Threshold $\alpha$}

To determine the optimal configuration for \algoname-Efficient, we conduct evaluations using various values of threshold $\alpha$.
Table~\ref{tab:varying_alpha} demonstrates that the performance of each threshold value is consistent across different model sizes.
Taking both prediction performance and efficiency into consideration, we have selected $\alpha=0.9$ as the default value for all experiments.
The stable choice of $\alpha$ ensures that \algoname-Efficient can automatically identify the optimal rank for new tasks and models without the need for extensive tuning, thereby minimizing the associated costs.

\begin{table}[htb!]
    \centering
    \caption{Varying threshold $\alpha$ in \algoname-Efficient.}
    \begin{tabular}{lccccccc}
        \toprule
        \textbf{Threshold $\alpha$} & \multicolumn{6}{c}{\textbf{GPT-small}} \\
        \cmidrule(lr){2-7}
        & \textbf{PPL} & \textbf{Rank} & \textbf{Runtime} & \textbf{Memory} & \textbf{Model Size} & \textbf{Params}\\
        \midrule
        Baseline & 19.3 & 768 & 23.9h & 27GB & 248Mb & 124M \\
        0.8 & 20.2 & 152 & 18.3h & 24.5GB & 163Mb & 81.8M \\
        0.85 & 19.9 & 192 & 18.8h & 24.6GB & 171Mb & 85.5M \\
        0.9 & 19.6 & 254 & 19.1h & 24.7GB & 182Mb & 91.2M \\
        \bottomrule
    \end{tabular}
    \vskip 0.1in
    \begin{tabular}{lccccccc}
        \toprule
        \textbf{Threshold $\alpha$} & \multicolumn{6}{c}{\textbf{GPT-small}} \\
        \cmidrule(lr){2-7}
        & \textbf{PPL} & \textbf{Rank} & \textbf{Runtime} & \textbf{Memory} & \textbf{Model Size} & \textbf{Params}\\
        \midrule
        Baseline & 20.6 & 1024 & 28.7h & 73GB & 1.4KMb & 354M & \\
        0.8 & 21.0 & 168 & 16.8h & 63.8GB & 778Mb & 194M & \\
        0.85 & 20.4 & 213 & 17.4h & 64.0GB & 823Mb & 205M & \\
        0.9 & 20.2 & 286 & 18.0h & 65.1GB & 895Mb & 223M & \\
        \bottomrule
    \end{tabular}
    \vskip 0.1in
    \begin{tabular}{lccccccc}
        \toprule
        \textbf{Threshold $\alpha$} & \multicolumn{6}{c}{\textbf{GPT-large}} \\
        \cmidrule(lr){2-7}
        & \textbf{PPL} & \textbf{Rank} & \textbf{Runtime} & \textbf{Memory} & \textbf{Model Size} & \textbf{Params} \\
        \midrule
        Baseline & 20.9 & 1280 & 55.6h & 71GB & 3.1KMb &774M& \\
        0.8 & 21.3 & 179 & 32.7h & 57.6GB & 1.5KMb &384M& \\
        0.85 & 20.8 & 230 & 33.7h & 58.0GB & 1.6KMb &408M& \\
        0.9 & 20.4 & 312 & 34.6h & 58.2GB & 1.8KMb &445M& \\
        \bottomrule
    \end{tabular}
    \label{tab:varying_alpha}
\end{table}

\section{Conclusion}
In this work, we generalize the Greedy Low-Rank Learning (GLRL) to arbitrary orthogonal initialization, leading to the development of Incremental Low-Rank Learning (\algoname). Our method is capable of discovering the intrinsic rank of networks and has demonstrated comparable performance to full-rank counterparts on training GPT-2, while utilizing a maximum of 33\% of total ranks throughout training. 
The efficient variant of \algoname also achieves a significant reduction of 37\% in total training time and 36\% in model size with a 10\% reduction of memory usage  when training GPT-medium on WikiText-103.

We believe our work offers a novel approach to training low-rank networks through automatic rank determination. In the future, we aim to expand our method to encompass various network architectures and datasets. Additionally, we intend to optimize our algorithm implementation to further improve its computational efficiency.

\section{Social Impact}

Our research stands to benefit society by potentially improving computational efficiency and reducing model size during deep learning training, potentially decreasing associated economic and environmental costs. 
In addition, our work may also help to improve the accessibility of deep learning to researchers and practitioners with limited computational resources.
\section*{Acknowledgments}
We thank Xinpeng Zhao and Jenny Wei for their helpful discussions. We gratefully thank NVIDIA for computational support. A. Anandkumar is supported in part by Bren endowed chair.

\bibliography{zotero,custom}
\bibliographystyle{unsrtnat}

\newpage
\appendix
\section{Proof}

In this section, we present the proof of our main analysis, as shown in Theorem~\ref{thm:network-difference}.

As introduced in the main text, we analyze the learning trajectory of a 3-layer linear network where $y = W^2 W^1 x$, $W^1 \in \R^{N_h \times N_x}$ and $W^2 \in \R^{N_y \times N_h}$ are the weight matrices of the first and second layers, respectively, and $N_h < N_x, N_y$. 

We assume the inputs are orthogonal, i.e., $x_i^T x_j = 0$ for $i \neq j$.
In this case, the continuous gradient flow follows the following differential equations:
\begin{equation}
\label{eq:evolution}
\frac{{\partial}}{{\partial t}} W^{1} = W^{2^T} \left( \Sigma^{yx} - W^{2} W^{1} \Sigma^{xx} \right), \quad\frac{{\partial}}{{\partial t}} W^{2} = \left( \Sigma^{yx} - W^{2} W^{1} \Sigma^{xx} \right) W^{21^T}.
\end{equation}

Since the inputs are orthogonal $\Sigma^{xx} = I$, the input-output correlation matrix $\Sigma^{yx}$ contains all information we need to learn the network. 
We decompose $\Sigma^{yx}$ using SVD as follows:
\begin{equation}
    \Sigma^{yx} = U^{yy} S^{yx} V^{xx^T} = \sum_{\alpha=1}^{N_x} s_\alpha u_\alpha v_\alpha^T.
\end{equation}
Learning the direction and strength of each mode $\alpha$ is crucial to interpolate the input-output correlation matrix $\Sigma^{yx}$.

To analyze the evolution of each mode independently, we let $a^\alpha$ be the $\alpha^{\text {th}}$ column of $\bar{W}^{1}$, and let $b^{\alpha T}$ be the $\alpha^{\text {th }}$ row of $\bar{W}^{2}$, where $W^{1}=\bar{W}^{1} V^{xx^T}, W^{2}=U^{yy} \bar{W}^{2}$. 
Based on Equation~\ref{eq:evolution} we can characterize the evolution of each mode using $a^\alpha$ and $b^\alpha$:
\begin{equation}
    \label{eq:evolution_mode}
    \frac{\partial}{\partial t} a^\alpha = \left(s_\alpha - a^\alpha \cdot b^\alpha\right) b^\alpha - \sum_{\gamma \neq \alpha} b^\gamma \left(a^\alpha \cdot b^\gamma\right), \quad \frac{\partial}{\partial t} b^\alpha = \left(s_\alpha - a^\alpha \cdot b^\alpha\right) a^\alpha - \sum_{\gamma \neq \alpha} a^\gamma \left(b^\alpha \cdot a^\gamma\right).
\end{equation}
For both $ \frac{\partial}{\partial t} a^\alpha$ and $\frac{\partial}{\partial t} b^\alpha$, the first term characterizes the cooperative learning of the strength $s_{\alpha}$ using the $a^\alpha$ and $b^\alpha$.
The second term characterizes the competitive learning of the direction $a^\alpha$ and $b^\alpha$ given the distraction from other directions $a^\gamma$ and $b^\gamma$.

It is difficult to solve Equation~\ref{eq:evolution_mode} given arbitrary weight initialization due to complex competitive interaction between modes.
Therefore, we assume the weight initialization follows $W^{2}_0=U^{yy} M^2 O^T, W^{1}_0=O M^1 V^{xx^T}$, where $M^2, M^1$ are diagonal matrices, and $O$ is an arbitrary orthogonal matrix. 
$W^{2}_0 W^{1}_0$ ensures that each mode $\alpha$ is learned independently right from the beginning of training, enabling us to analyze the learning trajectory of each mode separately.
$a^\alpha$ and $b^\alpha$ will remain parallel to a certain direction $r^\alpha$ throughout the learning process, and we can rewrite Equation~\ref{eq:evolution_mode} as follows:
\begin{equation}
    \frac{\partial}{\partial t} a =b(s-a b), \quad \frac{\partial}{\partial t} b =a(s-a b),
\end{equation}
where we let $a=a^\alpha \cdot r^\alpha, b=b^\alpha \cdot r^\alpha$, and $s=s^\alpha$.
By further assuming $a = b$ and $u = ab$, we obtain: 
\begin{equation}
    \frac{\partial}{\partial t} u=2 u(s-u) .
\end{equation}
Integrate the above equation to obtain:
\begin{equation}
    t=\tau \int_{u_0}^{u_f} \frac{d u}{2 u(s-u)}=\frac{\tau}{2 s} \ln \frac{u_f\left(s-u_0\right)}{u_0\left(s-u_f\right)},
\end{equation}
where $u_0$ is the initial value determined by $M^2$ and $M^1$, $u_f$ is the target value of strength, and $\tau$ is a constant. $t$ is the time it takes for $u$ to travel from $u_0$ to $u_f$. 

As we analyze the difference of the product matrix $D_t = A_t - A_0$, it is equivalent to analyzing the residual of each mode: $u_t - u_0$. To analyze the entire evolution ($u_f \approx s$) of $u$ over time, we yield the following equation:
\begin{equation}
    u_f(t)=\frac{s e^{2 s t / \tau}}{e^{2 s t / \tau}-1+s / u_0} - u_0.
\end{equation}
\section{Clarification on Greedy Low-Rank Learning}

In this section, we additionally clarify on the greedy low-rank learning hypothesis, which is presented in Theorem~\ref{thm:glrl}.

Several works have demonstrated the greedy low-rank learning behavior under various settings and assumptions.
\citet{li_towards_2021} prove it under matrix factorization setting for deep linear network by analyzing the asymptotic behavior of gradient flow under infinitesimal initialization.
\citet{jacot_saddle--saddle_2022} also demonstrate the saddle-to-saddle learning behavior for deep linear networks, although they prove the rank-one case only. 
\citet{razin_implicit_2021} further extend the discussion to the setting of tensor factorization.

A formal description of Theorem~\ref{thm:glrl} is given below:
\begin{theorem}
    \label{thm:glrl_formal}
    Let $\tilde{W}_r$ be the $r$-th critical point of a rank-$r$ subspace of $W$, and let $\tilde{W}_0 = 0$ be the saddle point at zero.
    From an infinitesimal initialization ($W_0 \approx \tilde{W}_0$), the gradient flow $G(W)$ first visits the critical point $\tilde{W}_1$. If $\tilde{W}_1$ is not a minimizer, $G(W)$ will expand the searching space to a rank-2 subspace and converge to the critical point $\tilde{W}_2$. If $\tilde{W}_2$ is also not a minimizer, this process continues until $G(W)$ reaches $\tilde{W}_{r^{*}}$ in a rank-$r^{*}$ subspace that minimizes the objective function, provided that $r^{*} < \rank(W)$.
\end{theorem}

The theorem implies the greedy low-rank learning trajectory, such that the gradient descent first searches over a rank-1 subspace of $A_{\theta}$, and then greedily increases the rank by one whenever it fails to reach the minimizer.

Proving this requires the analysis of the limiting flow $G_{r \to r+1}(W)$, which is the gradient flow between two critical points $\tilde{W}_r$ and $\tilde{W}_{r+1}$.
Theorem~\ref{thm:glrl_formal} holds by showing that the flows $G_{0 \to 1}(W)$, $G_{1 \to 2}(W)$, ..., $G_{{r^{*}-1} \to r^{*}}(W)$ all exist during learning, which is a general proving direction adopted by recent works.
The details of the proof can be found in \citet{li_towards_2021,jacot_saddle--saddle_2022}.

\section{\algoname-Efficient}

\begin{algorithm}[h!]
    \caption{\algoname-Efficient}
    \label{alg:algo-efficient}
    \begin{algorithmic}[1]
        \State \textbf{Rank determination:}
        \State Run \algoname (Algorithm~\ref{alg:incremental-learning}) for $T$ iterations and get the final rank $r^{l}_{T}$ for each layer $l$
        \State \textbf{Efficient training:}
        \State Reparametrize $W_t^{l} = U_t^{l} V_t^{l} + W_0^{l}$ where $U_t^{l} \in \sR^{p^{l} \times r^{l}_{T}}$, $V_t^{l} \in \sR^{r^{l}_{T} \times q^{l}}$, and $W_0^{l} \in \sR^{p^{l} \times q^{l}}$
        \State Compute the top $r^{l}$ singular vectors: $u^{l}, s^{l}, v^{l} \gets \mathrm{SVD}_{r^{l}}(W_t^{l})$
        \State $U_t^{l} \gets v^{l}$ and $V_t^{l} \gets u^{l}$
        \State Perform standard training over fixed low-rank network until convergence
    \end{algorithmic}
\end{algorithm}

We present \algoname-Efficient in Algorithm~\ref{alg:algo-efficient}.
We apply \algoname only in the first $T$ iterations of training to determine an appropriate rank for low-rank training.
In practice, we set $T$ to be the number of iterations of a single epoch, across all tasks. 
We find this is sufficiently large to let \algoname converge to an appropriate rank.
After determining the optimal rank $r^{*}$, we reparameterize the weights as a rank-$r^{*}$ factorization of $U^{*}V^{*}$ only, such that $W = U^{*}V^{*}$ where $U^{*}V^{*} = W_0 + UV$. 
This eliminates the need for dynamically representing a separated $W_0$, which reduces computational costs as it avoids additional matrix
multiplication.

We note that a separated $W_0$ is important for \algoname to determine layer ranks given $UV$.
Without $W_0$, the training suffers from vanishing gradients and fails to converge, as $U$ and $V$ are initialized to be extremely small in practice.

\section{Experiment Setup}
For different sizes of GPT-2 baseline, we directly use the standard implementations of GPT-2 from fly repo by Hazy Research \citep{dao_monarch_2022}, which leverages on Huggingface transformers library
and Nvidia’s Megatron-LM repo. We follow the training recipe of fly repo.

Note that we find GPT-2 medium and GPT-2 large is not stable for converging. To avoid diverging, we need to use double float precision 32 for training. For fair comparison, we fix float precision to be double 32 for InRank and InRank-Efficient on GPT-2 medium and GPT-2 large training as well. We use an effective batch size of 512, and use gradient accumulation to fit into available GPU memory. We report the configuration used in Table~\ref{tab:exp_configuration}.
\begin{table}[H]
    \centering
    \fontsize{7.6}{9}\selectfont
    \caption{Configuration of the GPT-2 experiments.}
    \setlength\tabcolsep{4pt}
    \begin{tabular}{llcccccc}
        \toprule
        \textbf{Model} & \textbf{Method} & \textbf{Optimizer} & \textbf{Weight Decay} & \textbf{Learning Rate} & \textbf{Precision} & \textbf{Warmup/Epoch} & \textbf{GPU Type} \\
        \midrule
        GPT-small & Baseline & AdamW & 0.1 & 6e-4 & 16 & 1/100 & V100 32GB \\
        & \algoname-Efficient & AdamW & 0.1 & 6e-4 & 16 & 1/100 & V100 32GB \\
        GPT-medium & Baseline & AdamW & 0.1 & 1.5e-4 & 32 & 1/100 & A100 80GB \\
        & \algoname-Efficient & AdamW & 0.1 & 1.5e-4 & 32 & 1/100 & A100 80GB  \\
        GPT-large & Baseline & AdamW & 0.1 & 1.5e-4 & 32 & 1/100 & A100 80GB \\
        & \algoname-Efficient & AdamW & 0.1 & 1.5e-4 & 32 & 1/100 & A100 80GB  \\
        \bottomrule
    \end{tabular}
    \label{tab:exp_configuration}
\end{table}

However, we find InRank and InRank-Efficient has the capability to regularize to ensure convergence even using single float precision 16 on GPT-2 medium and GPT-2 large training. We also report such results for InRank-Efficient with explained ratio 0.9 in Table~\ref{tab:single_precision_result}. As shown in Table~\ref{tab:single_precision_result}, with the benefit of regularization of InRank-Efficient, we can further reduce runtime by 27\%, memory usage by 38\% and model size by 50\% on GPT-medium training and runtime by 27\%, memory usage by 32\% and model size by 50\% on GPT-large training, both with comparable prediction performance, compared with training under the double float precision 32.
\begin{table}[H]
    \centering
    \fontsize{9}{9}\selectfont
    \caption{Performance comparison of double and single float precision.}
    \setlength\tabcolsep{4pt}
    \begin{tabular}{lccccccc}
        \toprule
        \textbf{Model} & \textbf{Precision} & \textbf{PPL} & \textbf{Rank} & \textbf{Runtime} & \textbf{Memory} & \textbf{Model Size} & \textbf{Params} \\
        \midrule
        \algoname-Efficient-GPT-medium & 32 & \textbf{20.2} & 286 & 18.0h & 65GB & 895Mb & 223M \\
                   & 16 & 20.3 & 286 & \textbf{13.2h} & \textbf{40GB} &\textbf{447Mb} & 223M  \\
        \algoname-Efficient-GPT-large & 32 & \textbf{20.4} & 312 & 34.6h & 58GB & 1.8KMb & 445M \\
                  & 16 & 20.5 & 312 & \textbf{25.1h} &\textbf{39GB}& \textbf{891Mb} & 445M  \\
        \bottomrule
    \end{tabular}
    \label{tab:single_precision_result}
\end{table}

\section{Rank Evolution during Training}
\begin{figure}[H]
    \centering
    \includegraphics[width=1.0\linewidth]{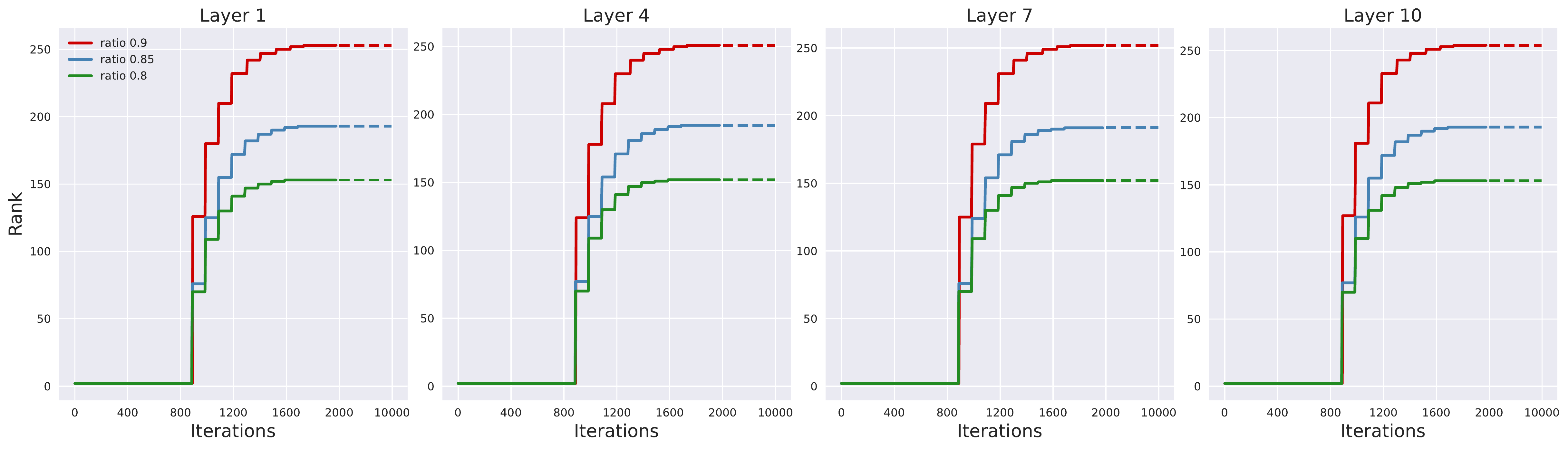}
    \caption{The rank evolution in various MLP layers when applying \algoname on GPT-small model.}
    \label{fig:rank-evolution1}
\end{figure}

\begin{figure}[H]
    \centering
    \includegraphics[width=1.0\linewidth]{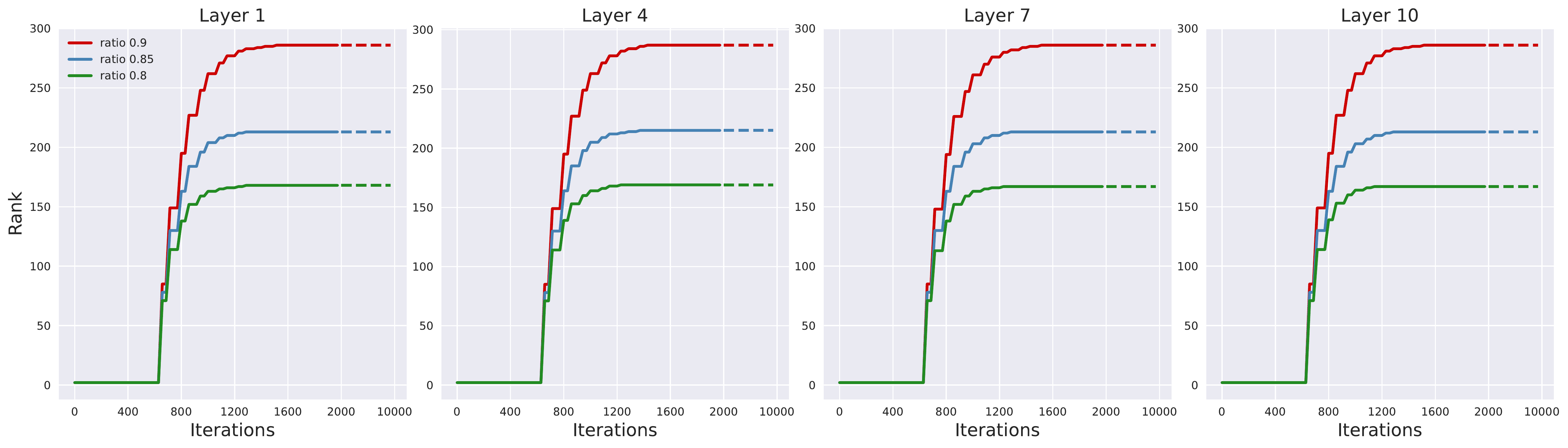} 
    \includegraphics[width=1.0\linewidth]{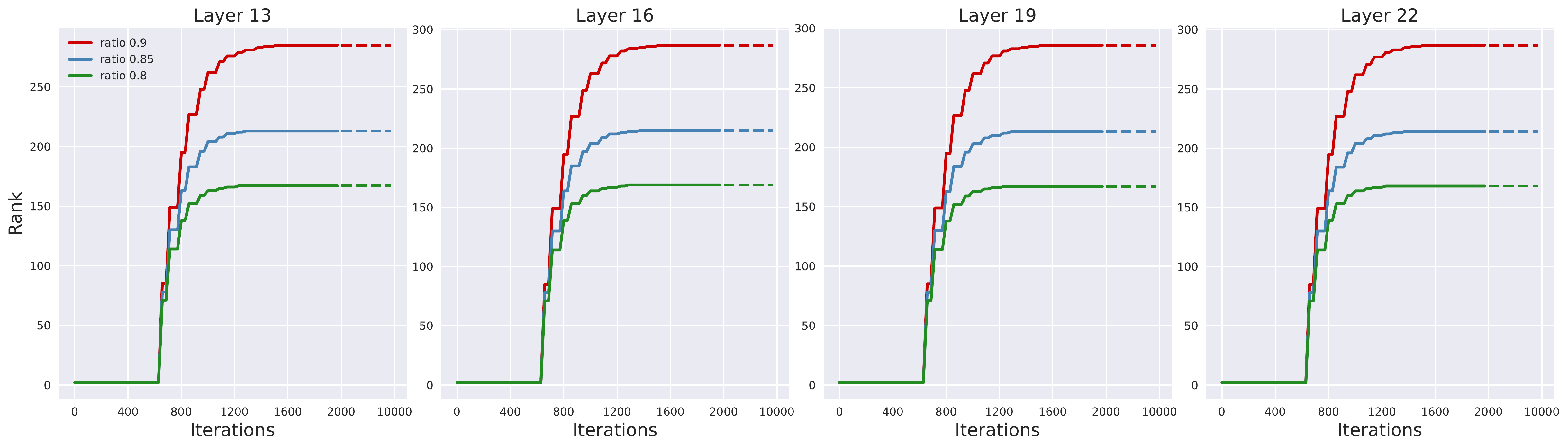}
    \caption{The rank evolution in various MLP layers when applying \algoname on GPT-medium model.}
    \label{fig:rank-evolution2}
\end{figure}

\begin{figure}[H]
    \centering
    \includegraphics[width=1.0\linewidth]{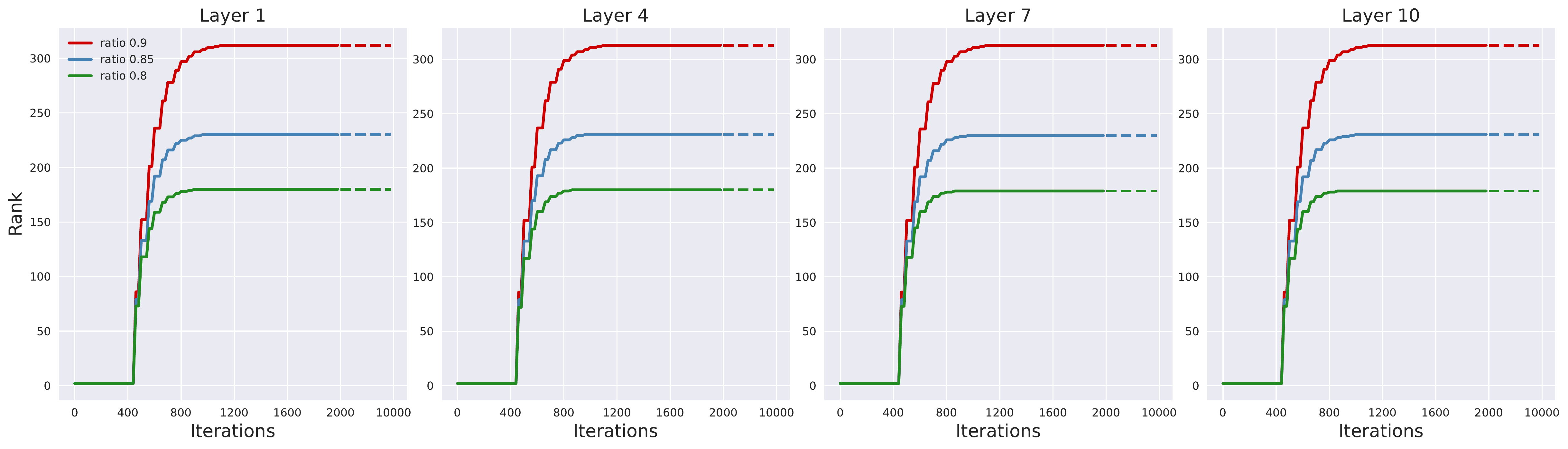} 
    \includegraphics[width=1.0\linewidth]{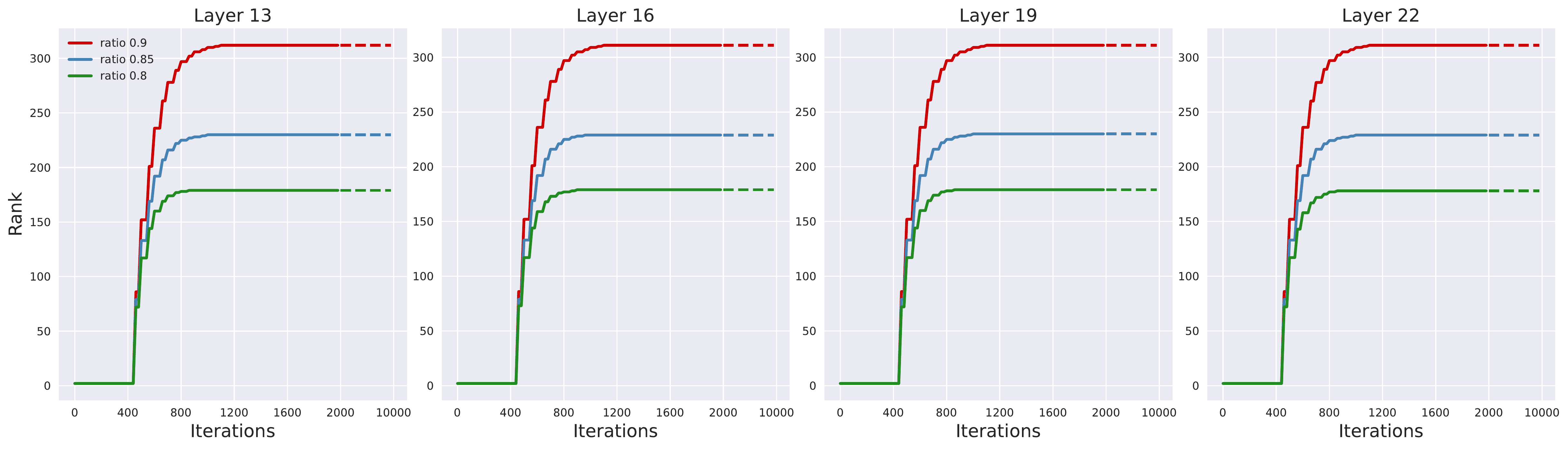}
    \includegraphics[width=1.0\linewidth]{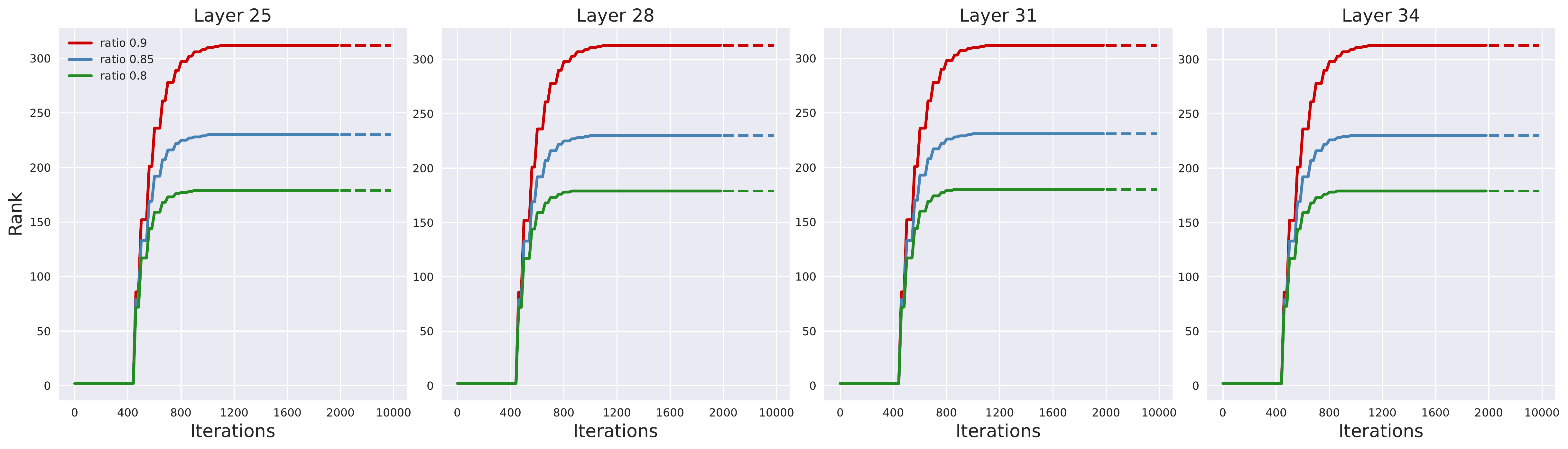}
    \caption{The rank evolution in various MLP layers when applying \algoname on GPT-large model.}
    \label{fig:rank-evolution3}
\end{figure}

We present the rank evolution in various MLP layers when applying \algoname on different sizes of GPT models. 
As shown in Figure~\ref{fig:rank-evolution1}, Figure~\ref{fig:rank-evolution2}, and Figure~\ref{fig:rank-evolution3}, we visualize the rank evolution over the first 5\% of the total training iterations.
The figures indicate that the increment of rank mostly happens in the early stage of training.


\section{Low-Rank Learning in Practice}

In this section, we provide additional results demonstrating that the cumulative weight updates follow the low-rank learning trajectory over a broad 
range of network architectures and learning algorithms.

\subsection{Low-Rank learning under different architectures}
\subsubsection{RNN}
\begin{figure}[H]
    \centering
    \includegraphics[width=1.0\linewidth]{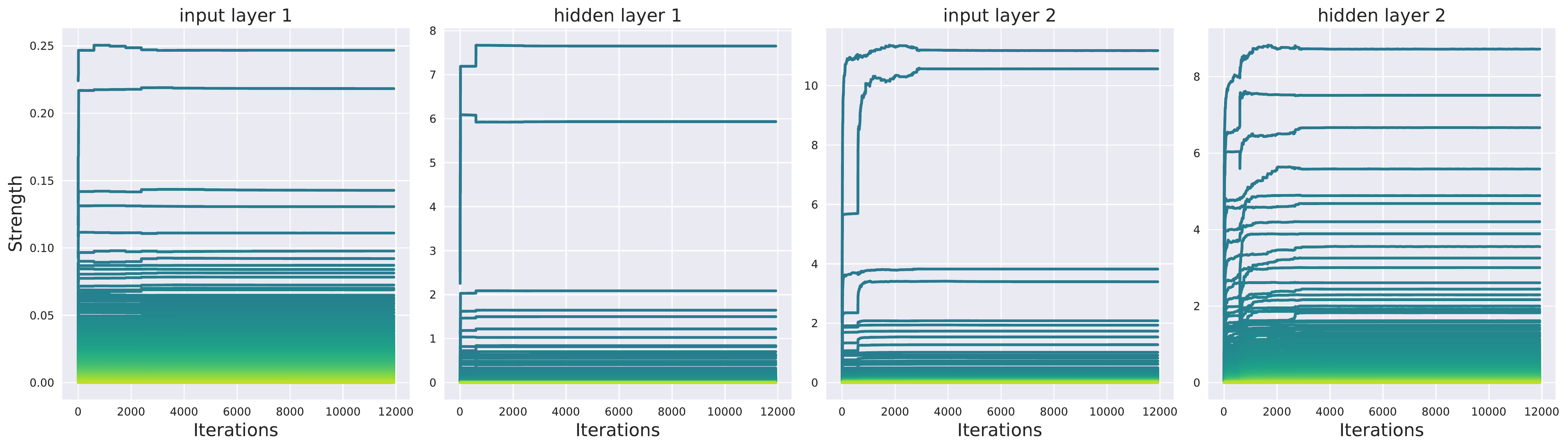}
    \includegraphics[width=0.5\linewidth]{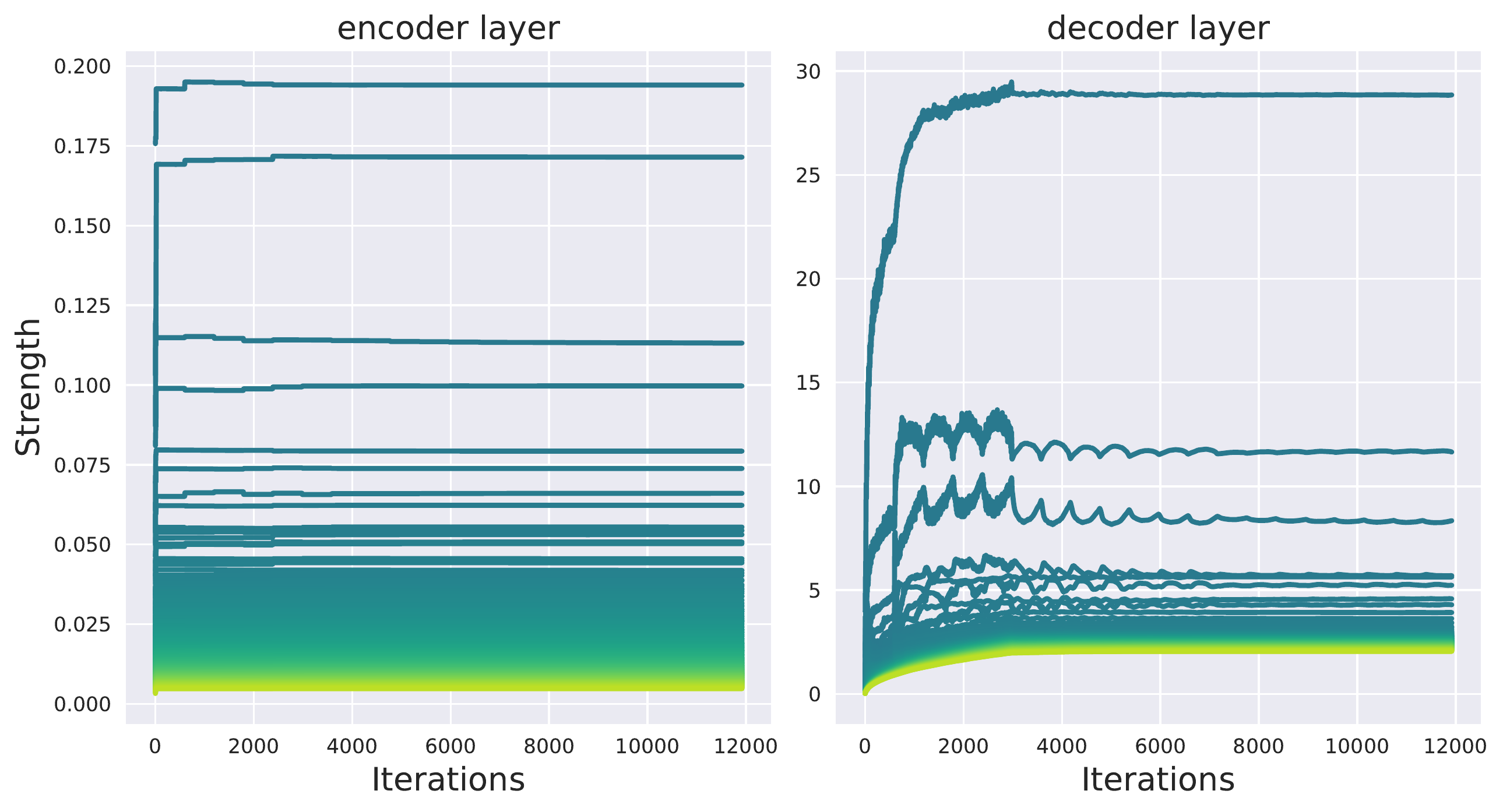}
    \caption{The evolutions of all singular vectors of cumulative weight updates $D_t$ over the training of RNN.}
\end{figure}
\subsubsection{GRU}
\begin{figure}[H]
    \centering
    \includegraphics[width=0.75\linewidth]{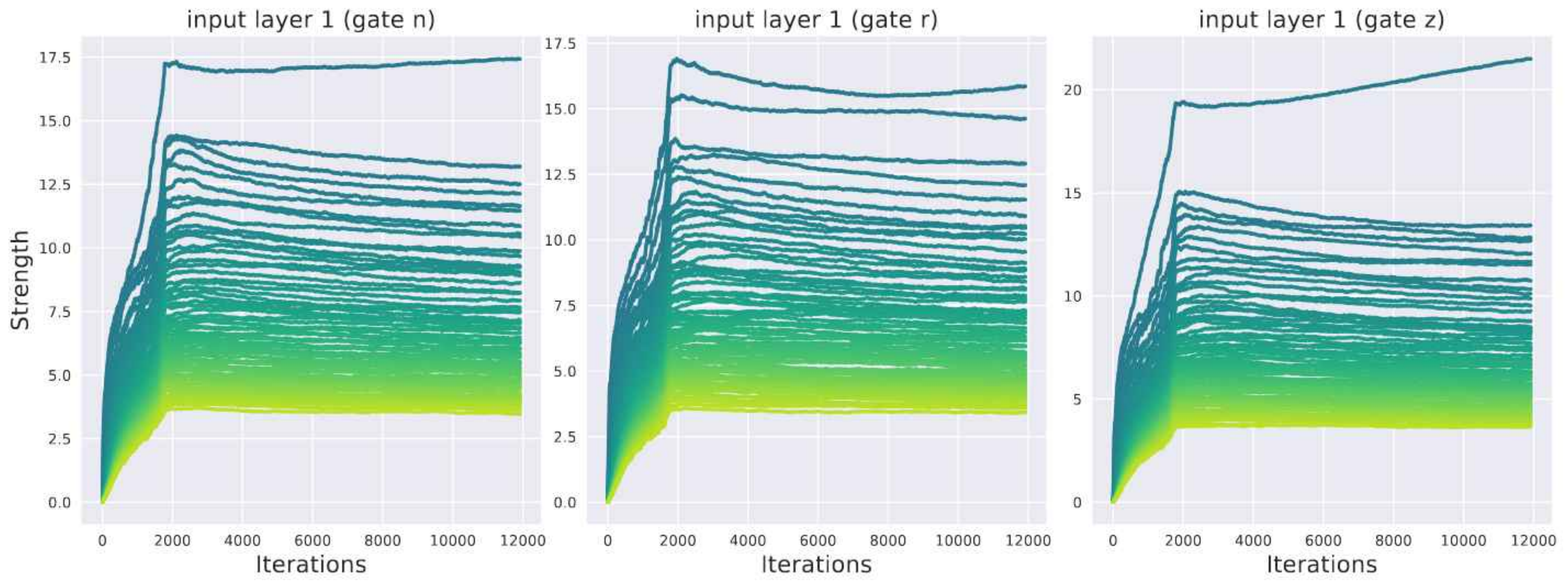}
    \includegraphics[width=0.75\linewidth]{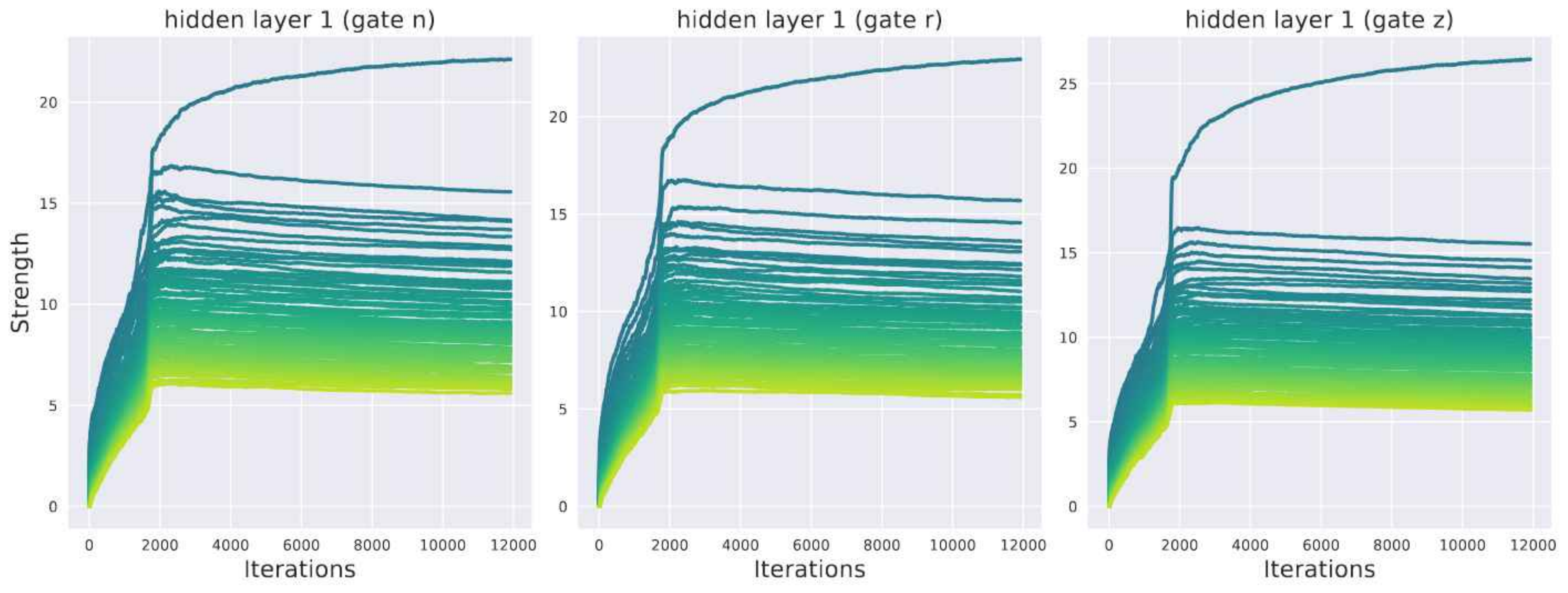}
\end{figure}
\begin{figure}[H]
    \centering
    \includegraphics[width=0.75\linewidth]{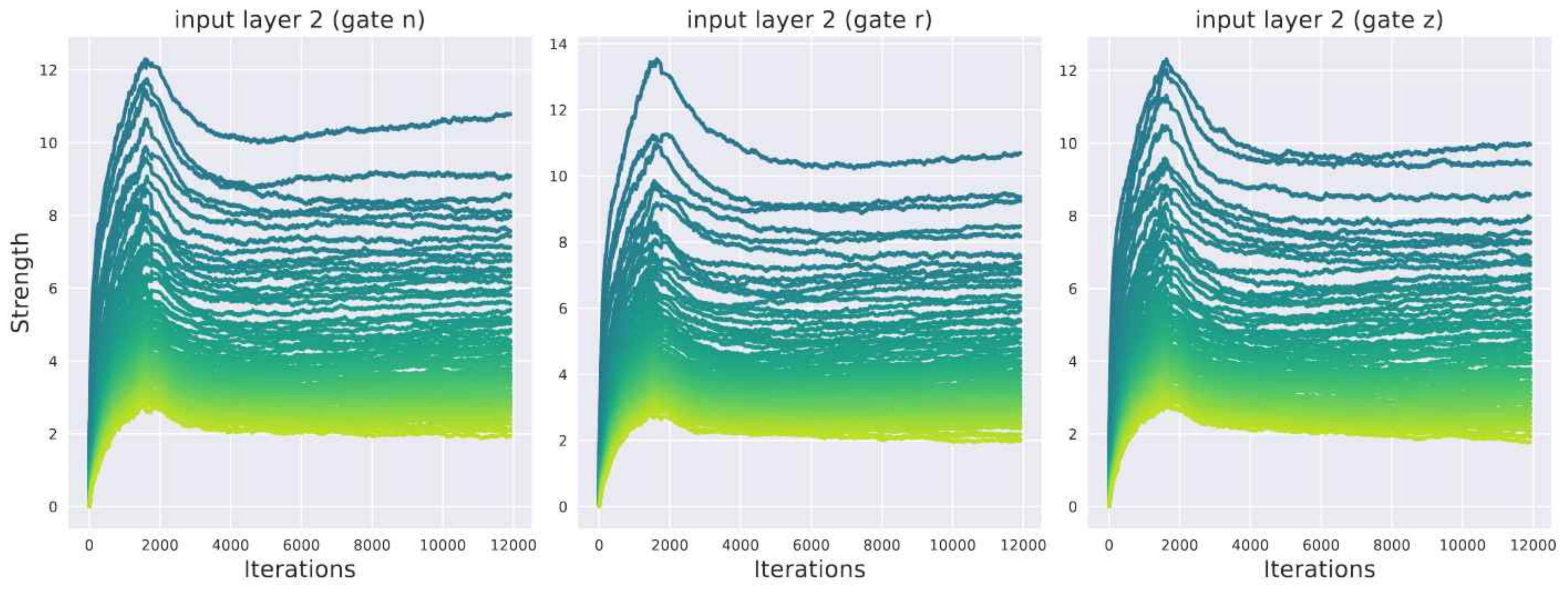}
        \includegraphics[width=0.75\linewidth]{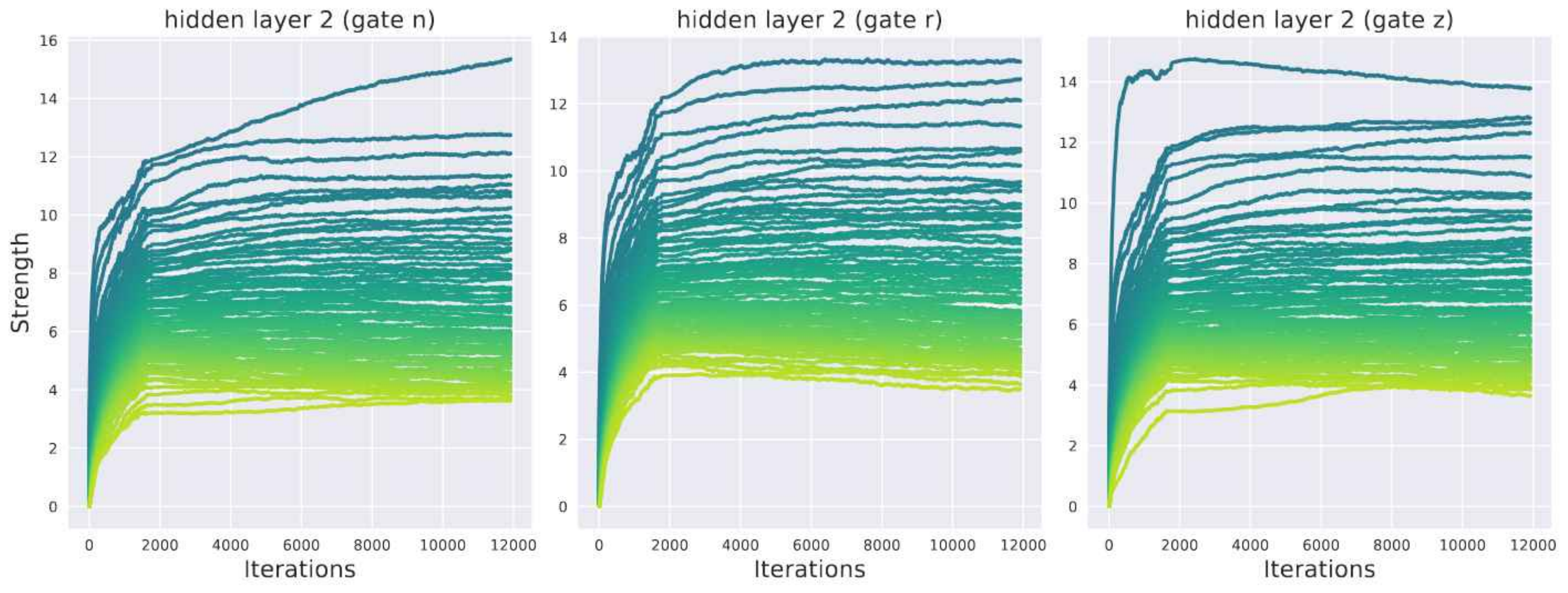}
    \includegraphics[width=0.25\linewidth]{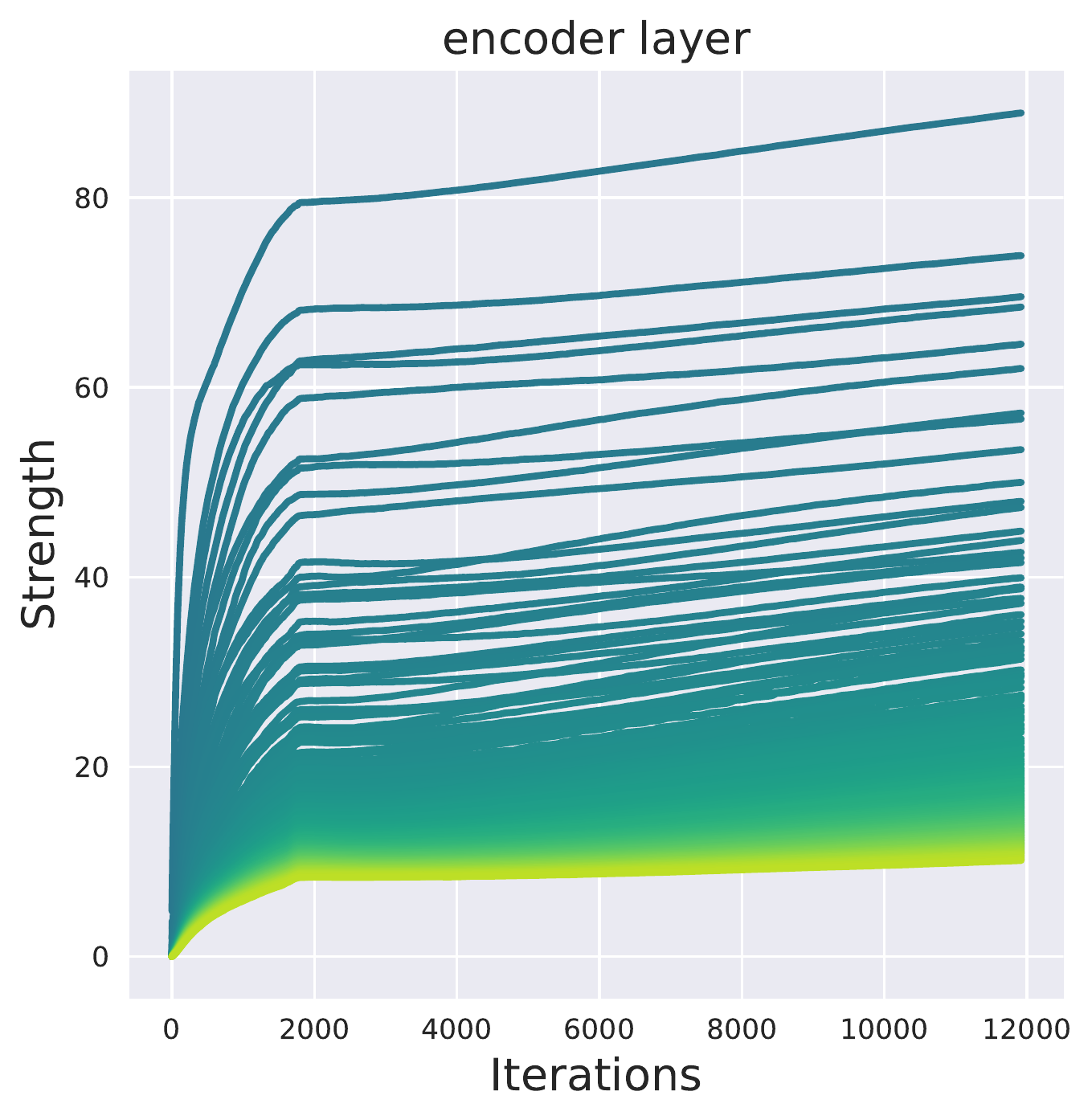}
    \caption{The evolutions of all singular vectors of cumulative weight updates $D_t$ over the training of GRU.}
\end{figure}
\subsubsection{LSTM}
\begin{figure}[H]
    \centering
    \includegraphics[width=1.0\linewidth]{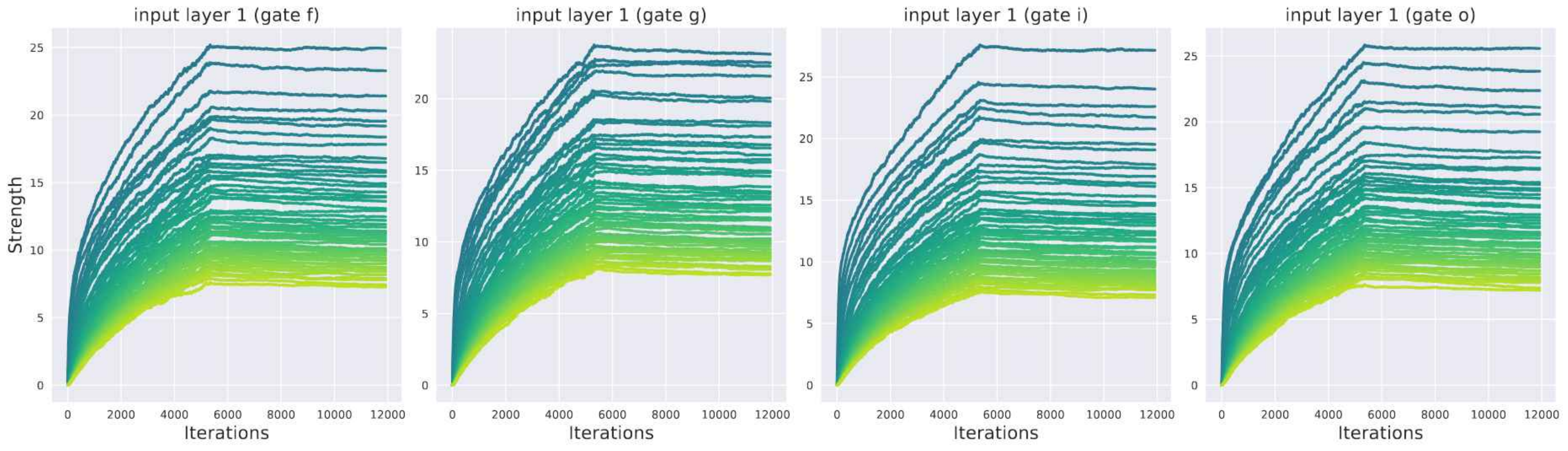}
    \includegraphics[width=1.0\linewidth]{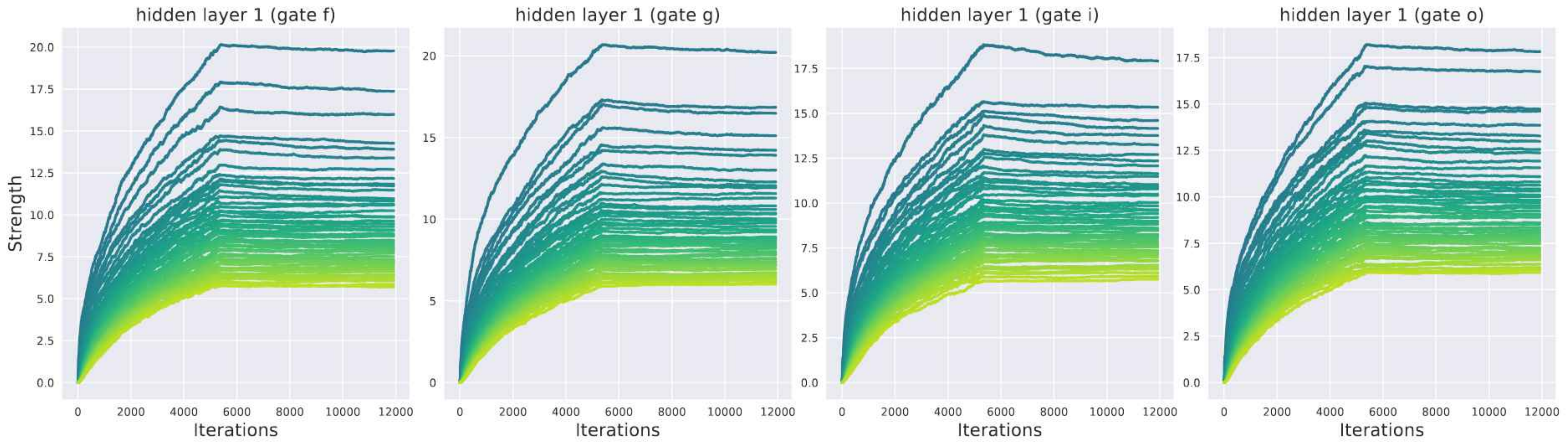}
\end{figure}
\begin{figure}[H]
    \centering
    \includegraphics[width=1.0\linewidth]{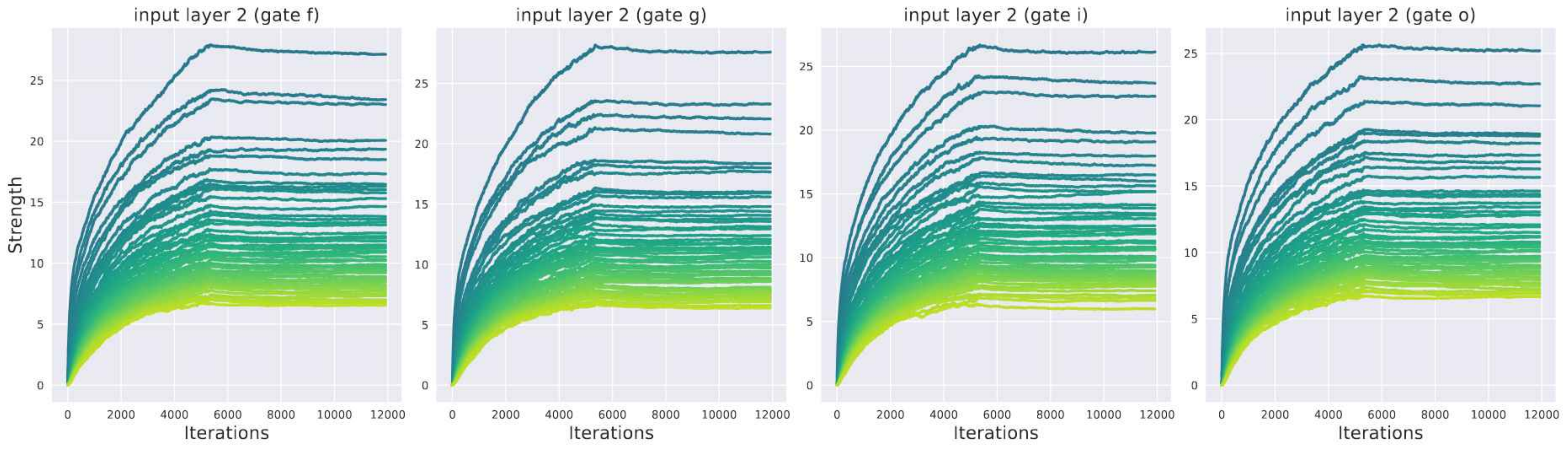}
    \includegraphics[width=1.0\linewidth]{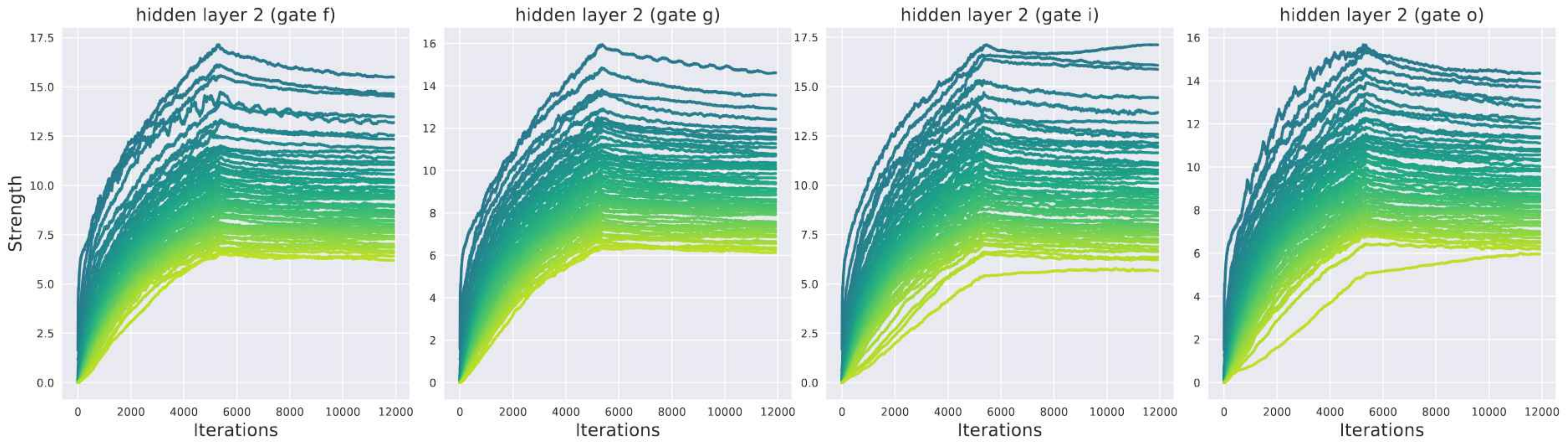}
    \includegraphics[width=0.25\linewidth]{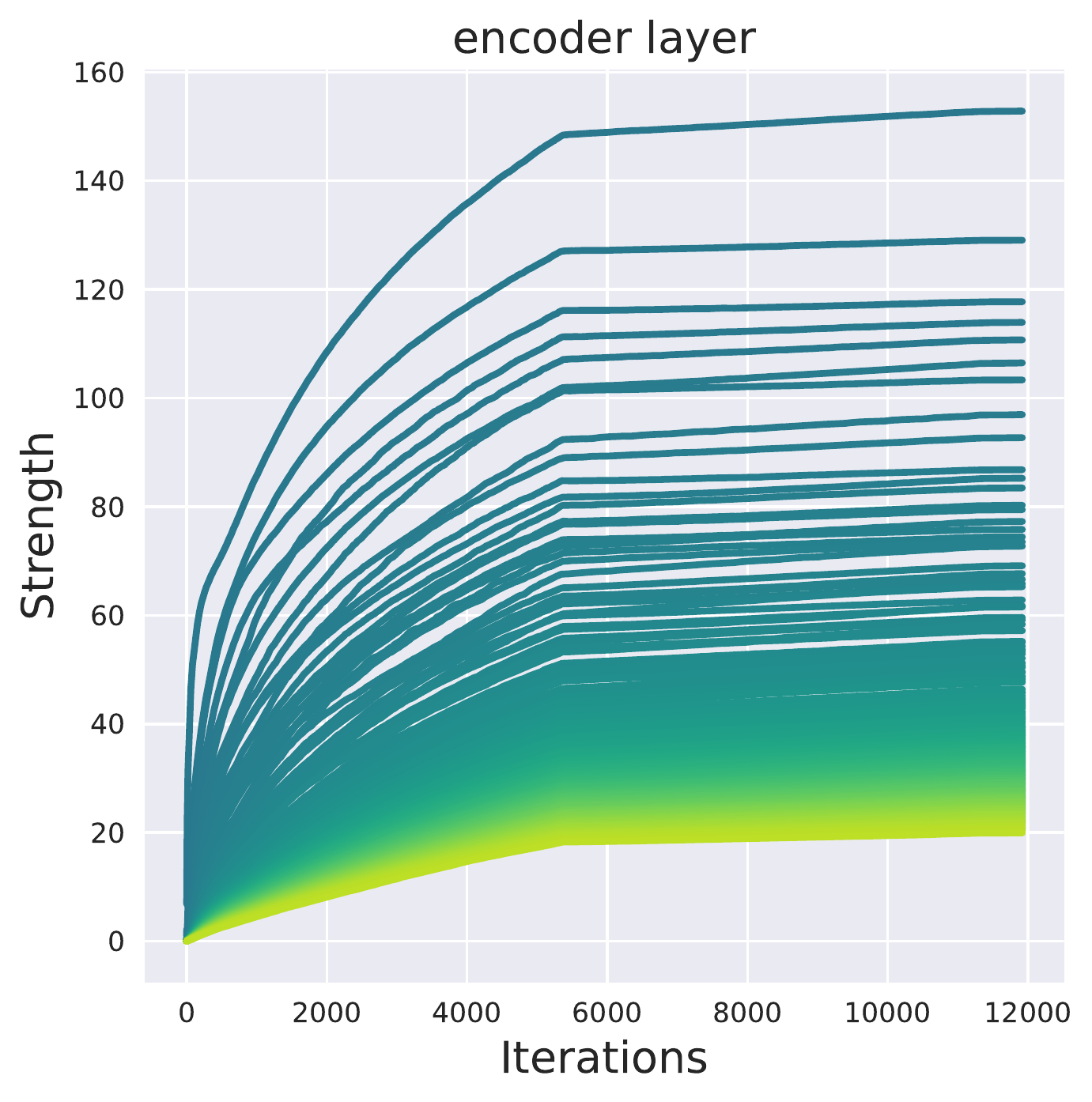}
    \caption{The evolutions of all singular vectors of cumulative weight updates $D_t$ over the training of LSTM.}
\end{figure}

\subsubsection{Transformer}
\begin{figure}[H]
    \centering
    \includegraphics[width=0.75\linewidth]{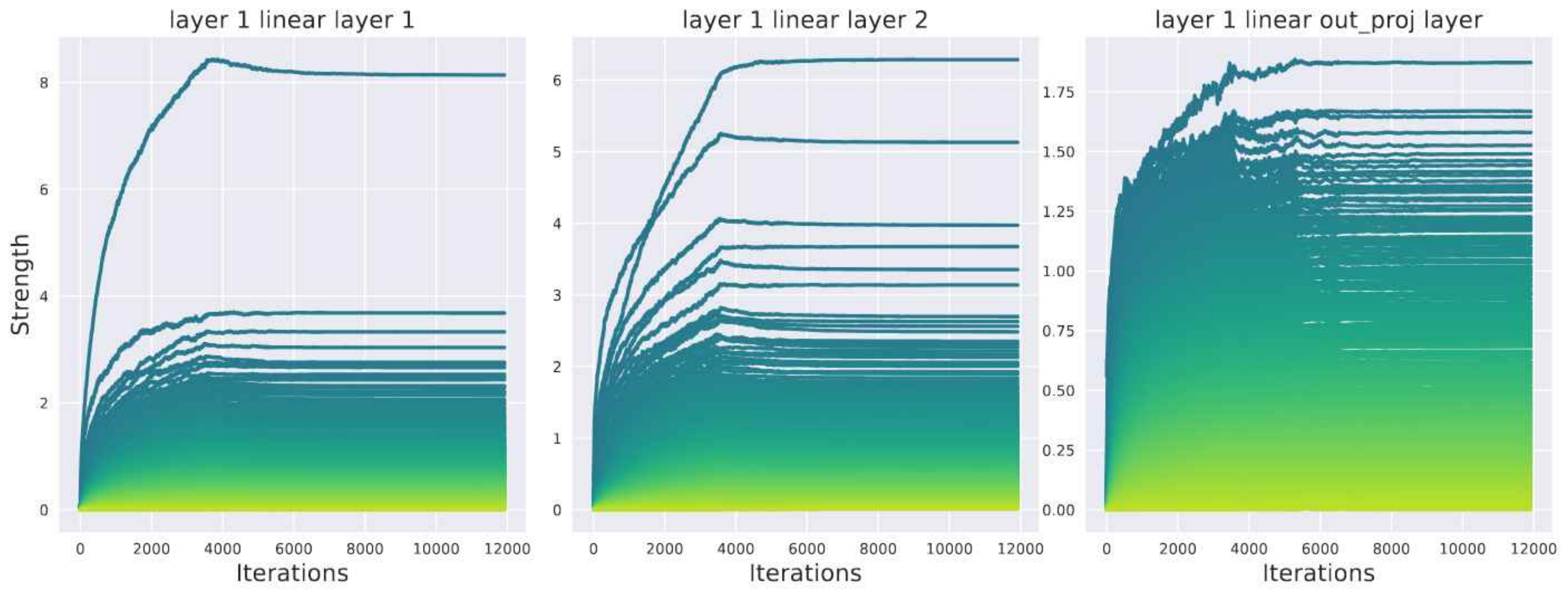}
    \includegraphics[width=0.75\linewidth]{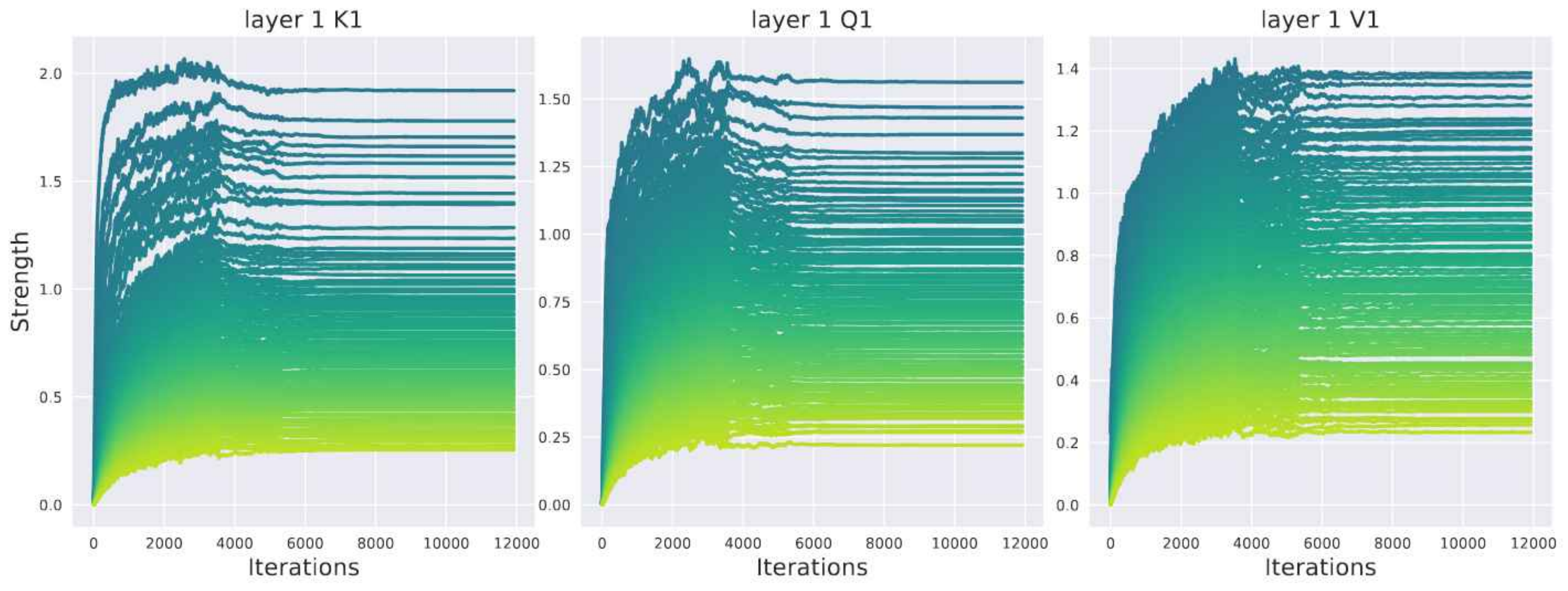}
\end{figure}
\begin{figure}[H]
    \centering
    \includegraphics[width=0.75\linewidth]{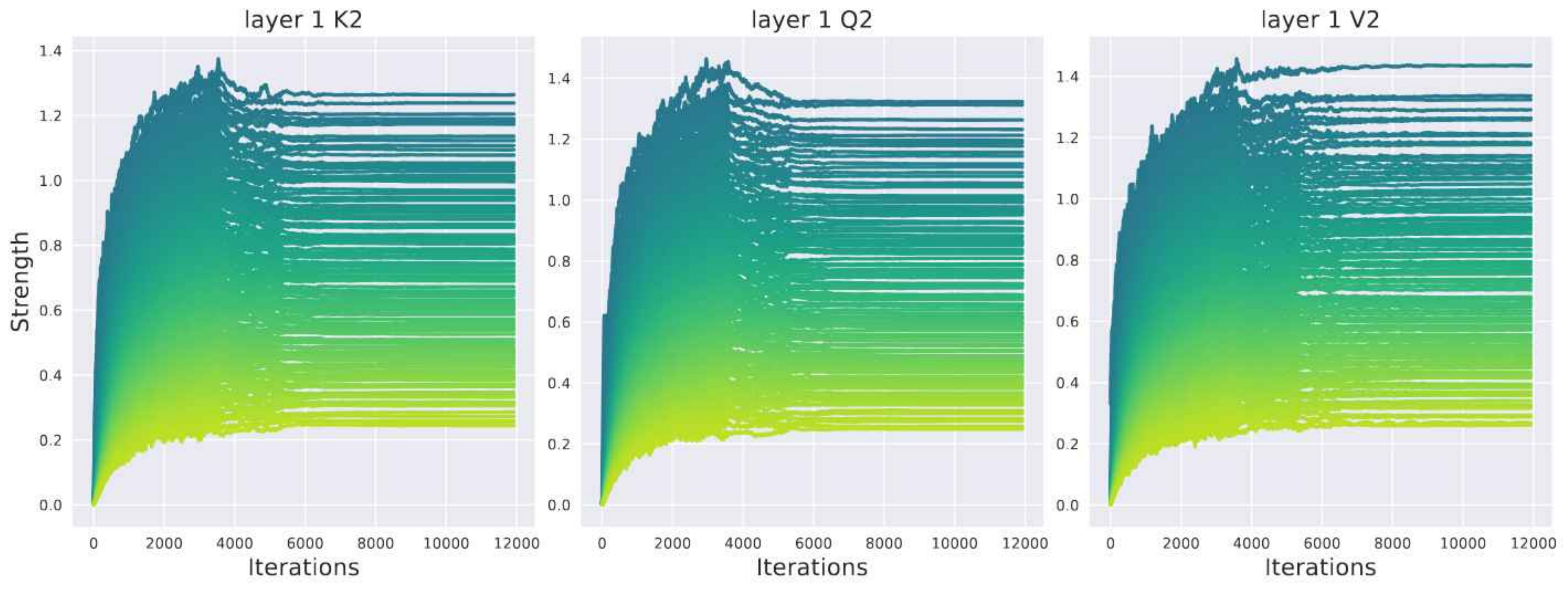}
    \includegraphics[width=0.75\linewidth]{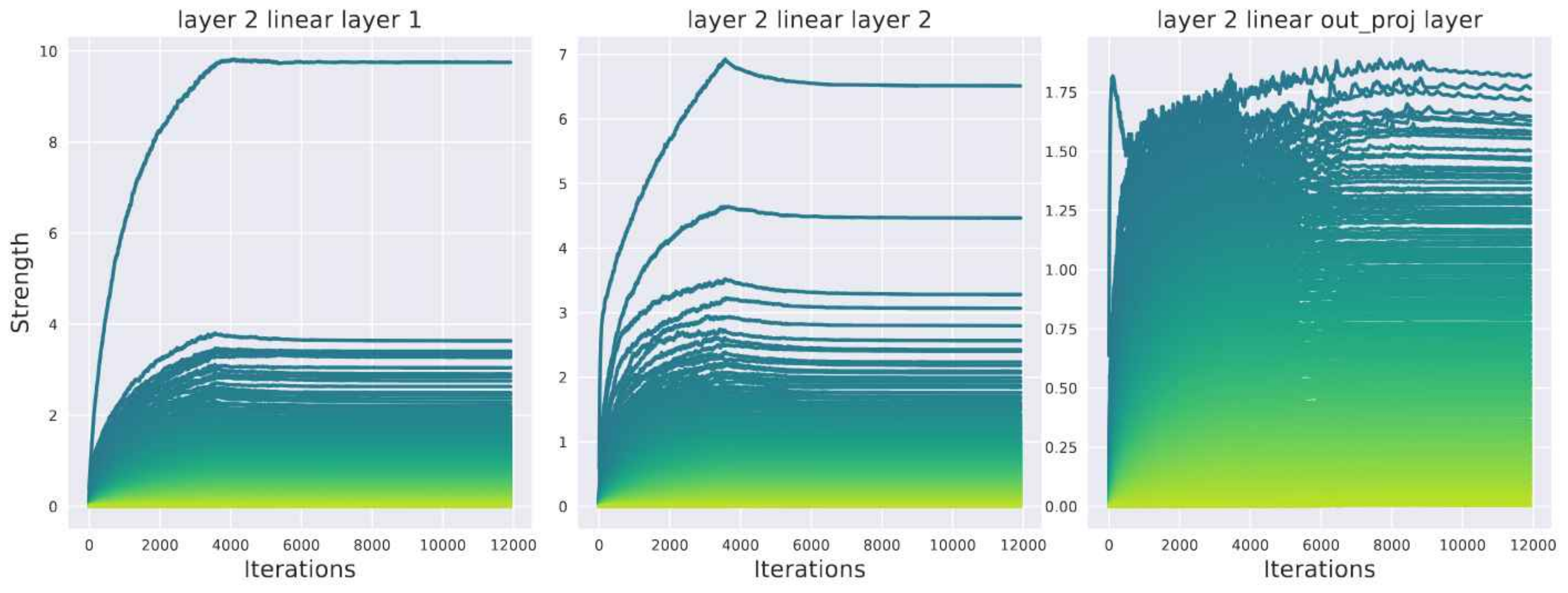}
    \includegraphics[width=0.75\linewidth]{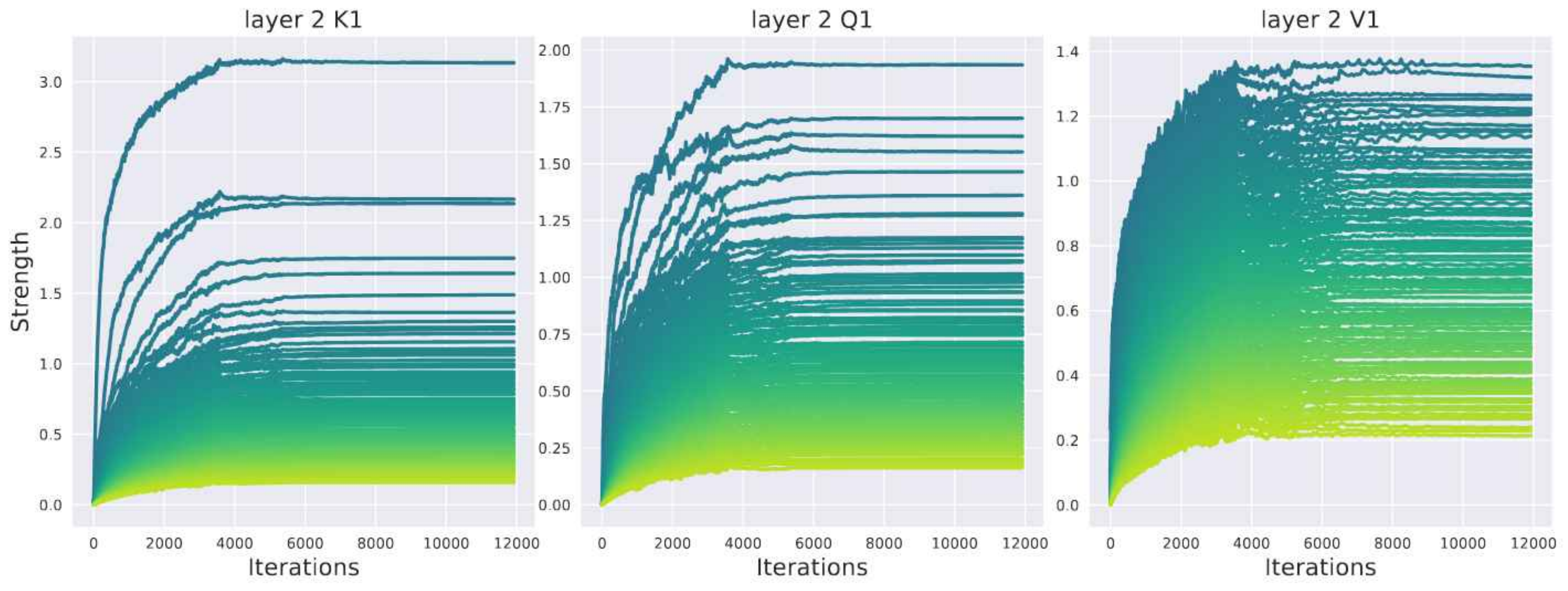}
    \includegraphics[width=0.75\linewidth]{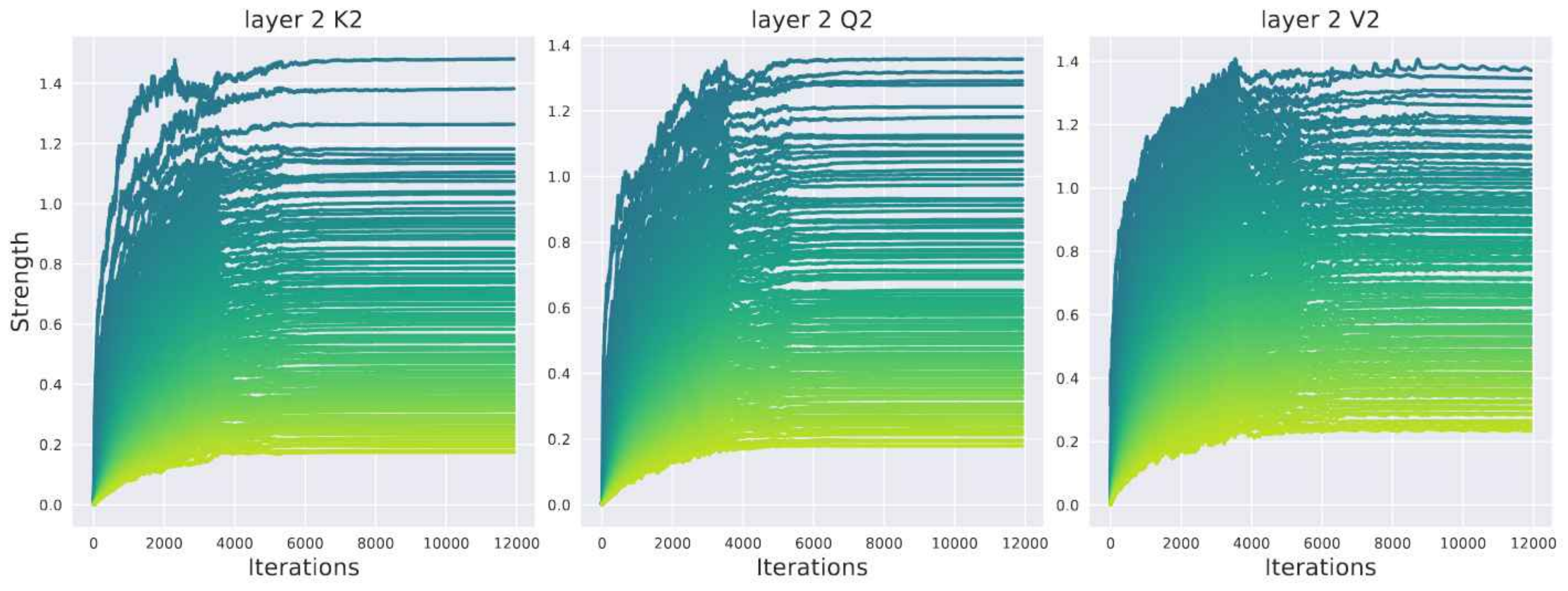}
    \includegraphics[width=0.5\linewidth]{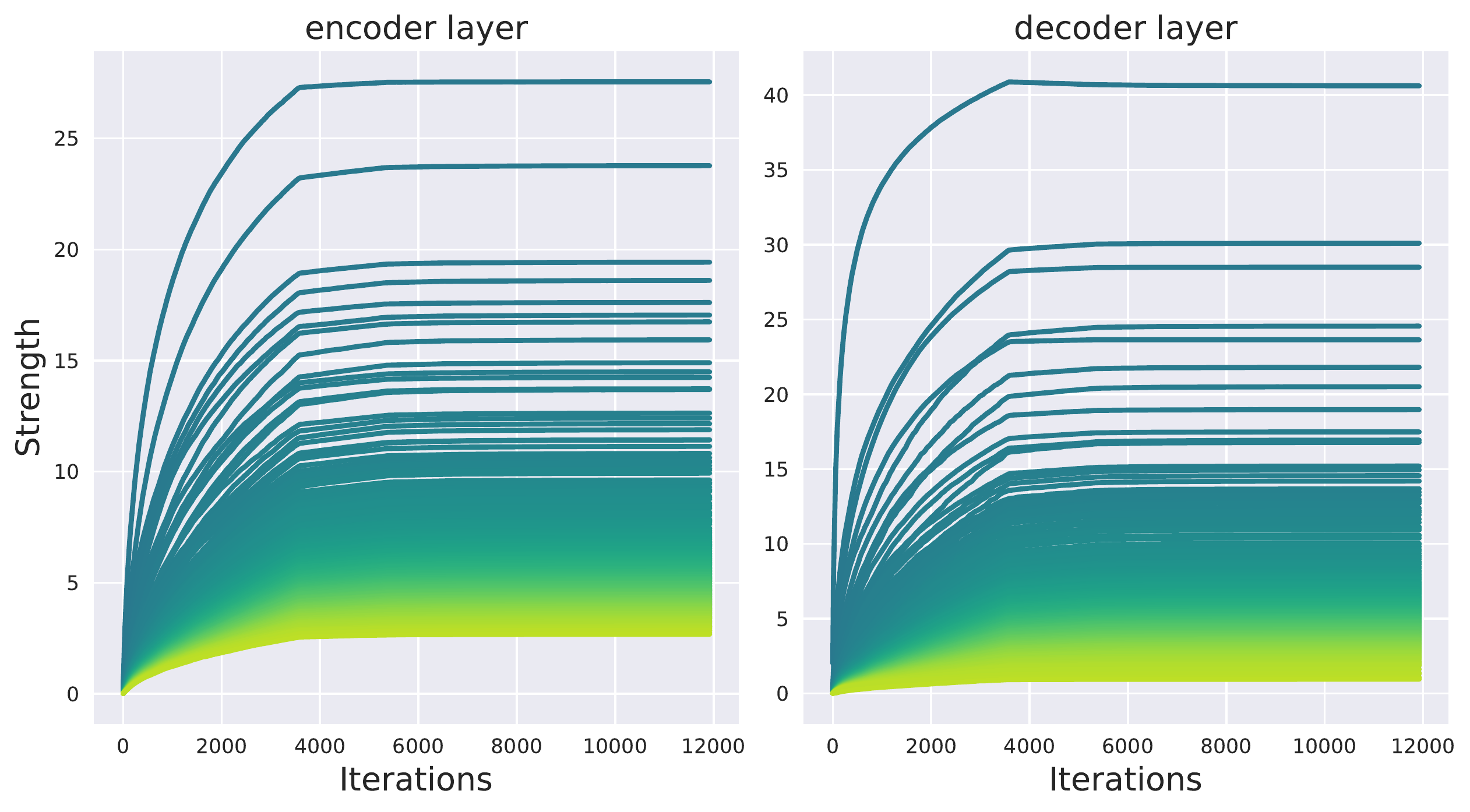}
    \caption{The evolutions of all singular vectors of cumulative weight updates $D_t$ over the training of Transformer.}
\end{figure}
Compared among these architectures, we find that high complexity of activation function inhibits the low-rank bias in the cumulative weight updates. For example, clearly, GRU, LSTM and attention-layer in Transformer triggers all singular values up above $0$, while RNN and feedforward layer in Transformer, the complexity of activation function more similar to MLP layer with ReLU shown in Figure~\ref{fig:spectrum_over_training}, behave the low-rank bias clearly in the cumulative weight updates.

\subsection{Being deeper encourages low-rank bias}
\begin{figure}[H]
    \centering
    \includegraphics[width=0.25\linewidth]{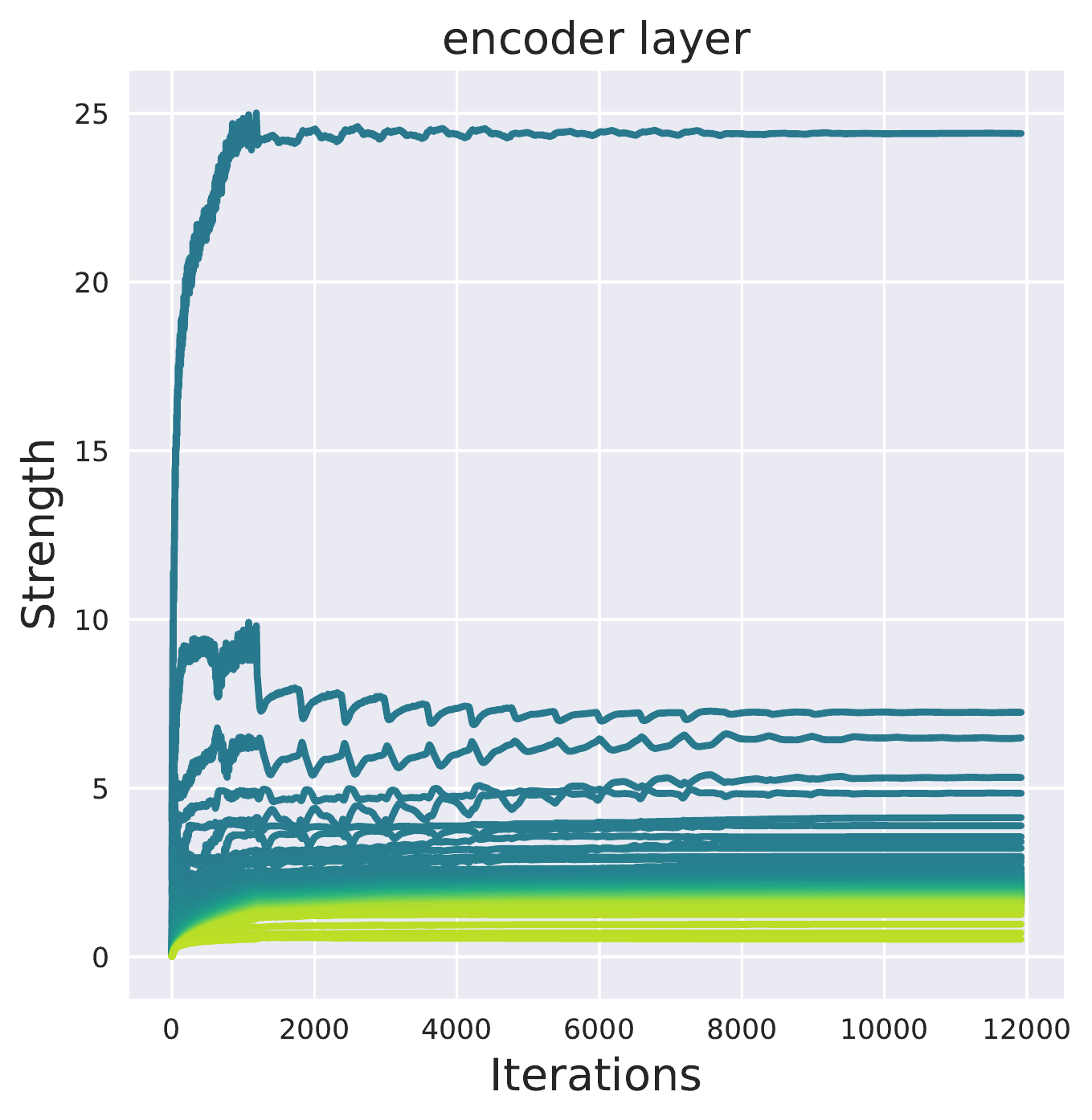}
    \caption{The evolutions of all singular vectors of cumulative weight updates $D_t$ over the training of 4-layer GRU. To save space, we select layer 1 and layer 3 to display.}
    \label{fig:spectrum_over_training_GRU}
\end{figure}

\begin{figure}[H]
    \centering
    \includegraphics[width=1.0\linewidth]{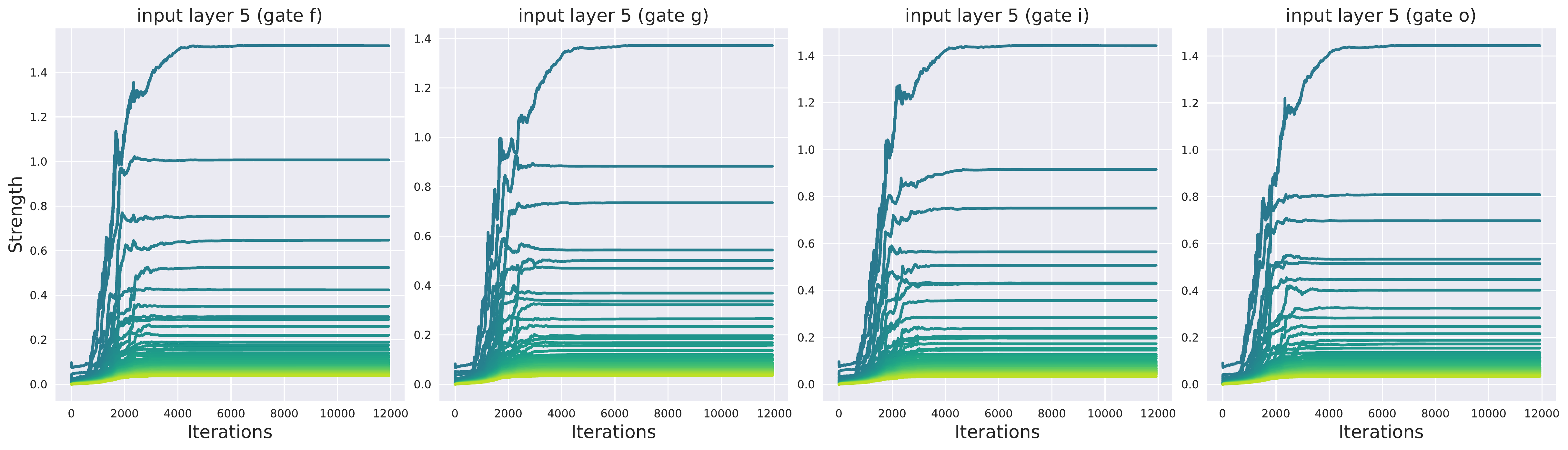}
    \includegraphics[width=1.0\linewidth]{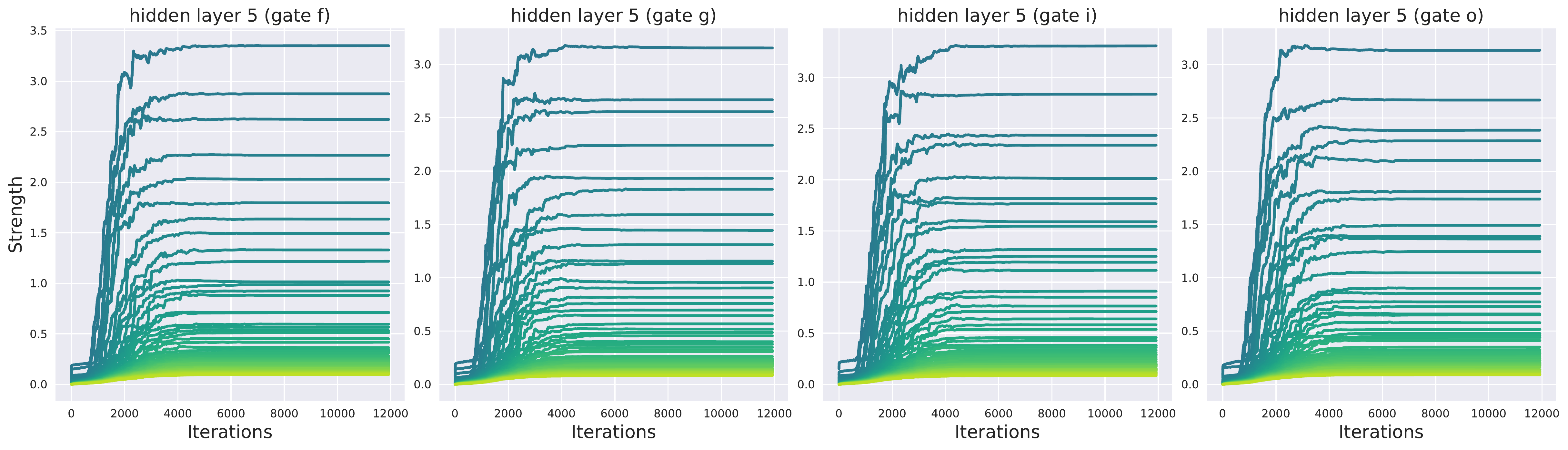}
    \includegraphics[width=0.25\linewidth]{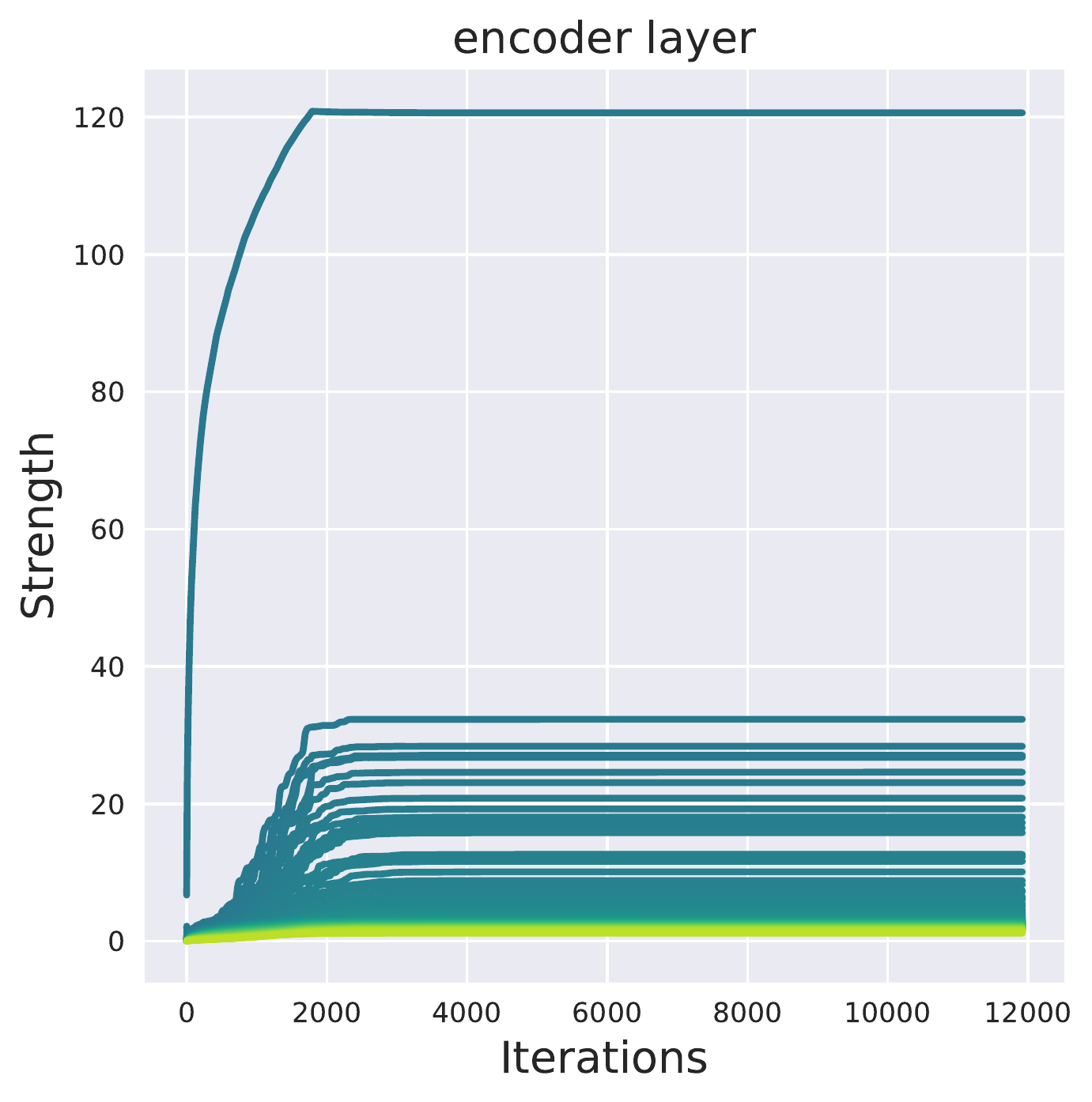}
    \caption{The evolutions of all singular vectors of cumulative weight updates $D_t$ over the training of 6-layer LSTM. To save space, we select layer 1, layer 3, and layer 5 to display.}
    \label{fig:spectrum_over_training_LSTM}
\end{figure}

Consistent with the results from the works of \citet{arora_implicit_2019} and \citet{li_towards_2021}, we also find being deeper encourages low-rank bias. As demonstrated in Figure~\ref{fig:spectrum_over_training_GRU} and Figure~\ref{fig:spectrum_over_training_LSTM}, when we remain other hyperparameters the same but add GRU from two layers to four layers and LSTM from two layers to six layers, GRU and LSTM start to behave the low-rank bias clearly. Because LSTM has higher complexity of activation function than GRU, LSTM needs more layers to start to behave the low-rank bias clearly. Similarly, attention layer in Transformer needs more layers than LSTM because attention layer in Transformer has higher complexity of activation function than LSTM.
\subsubsection{Four depth GRU}
\begin{figure}[H]
    \centering
    \includegraphics[width=0.75\linewidth]{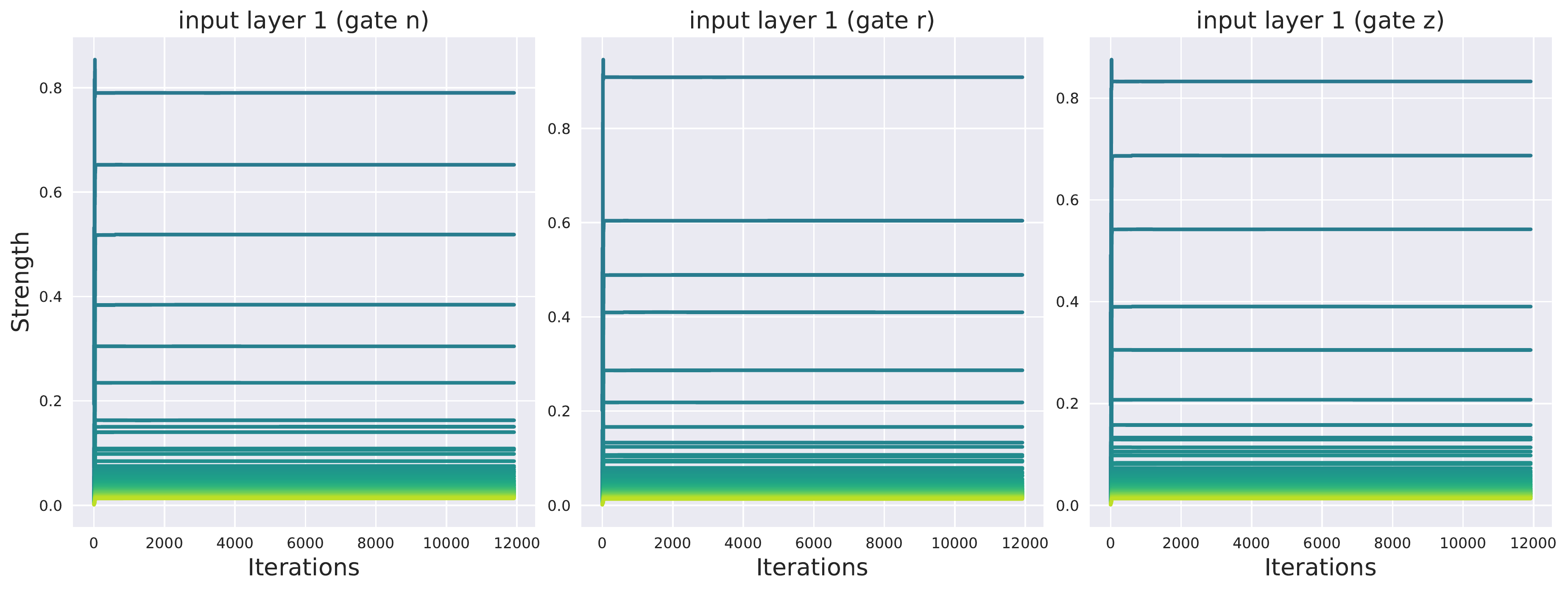}
    \includegraphics[width=0.75\linewidth]{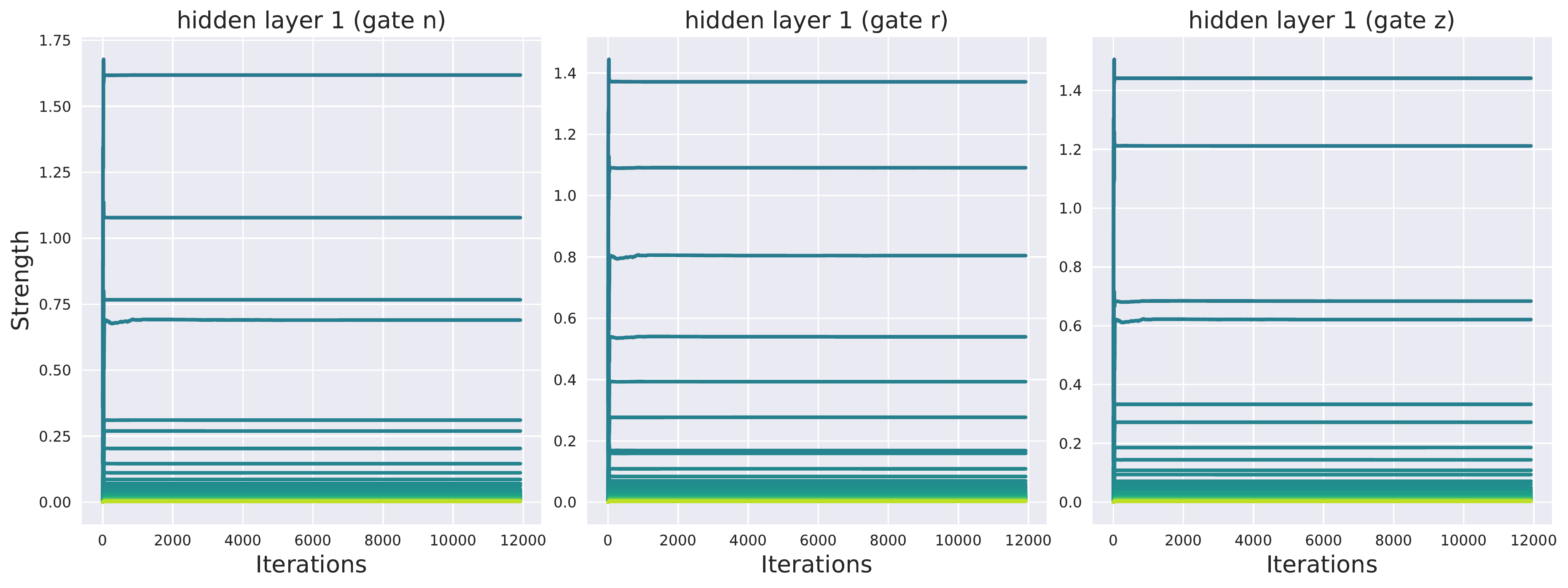}
    \includegraphics[width=0.75\linewidth]{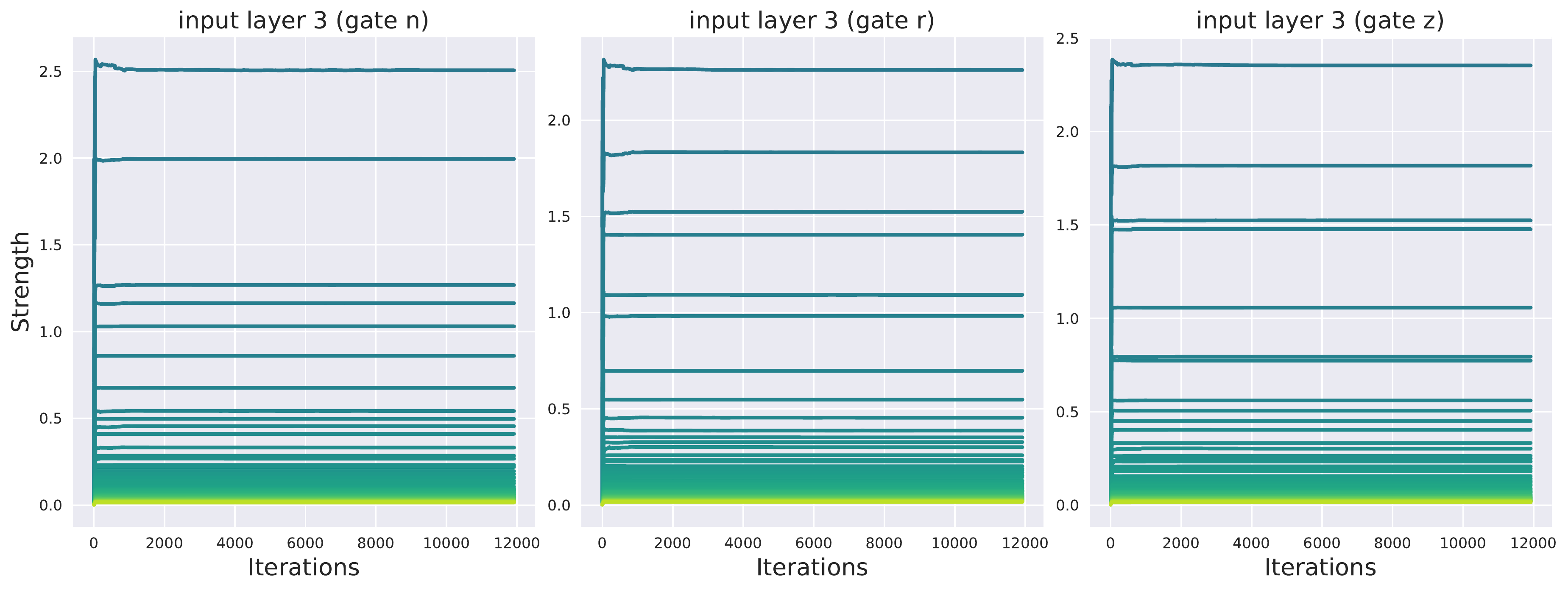}
    \includegraphics[width=0.75\linewidth]{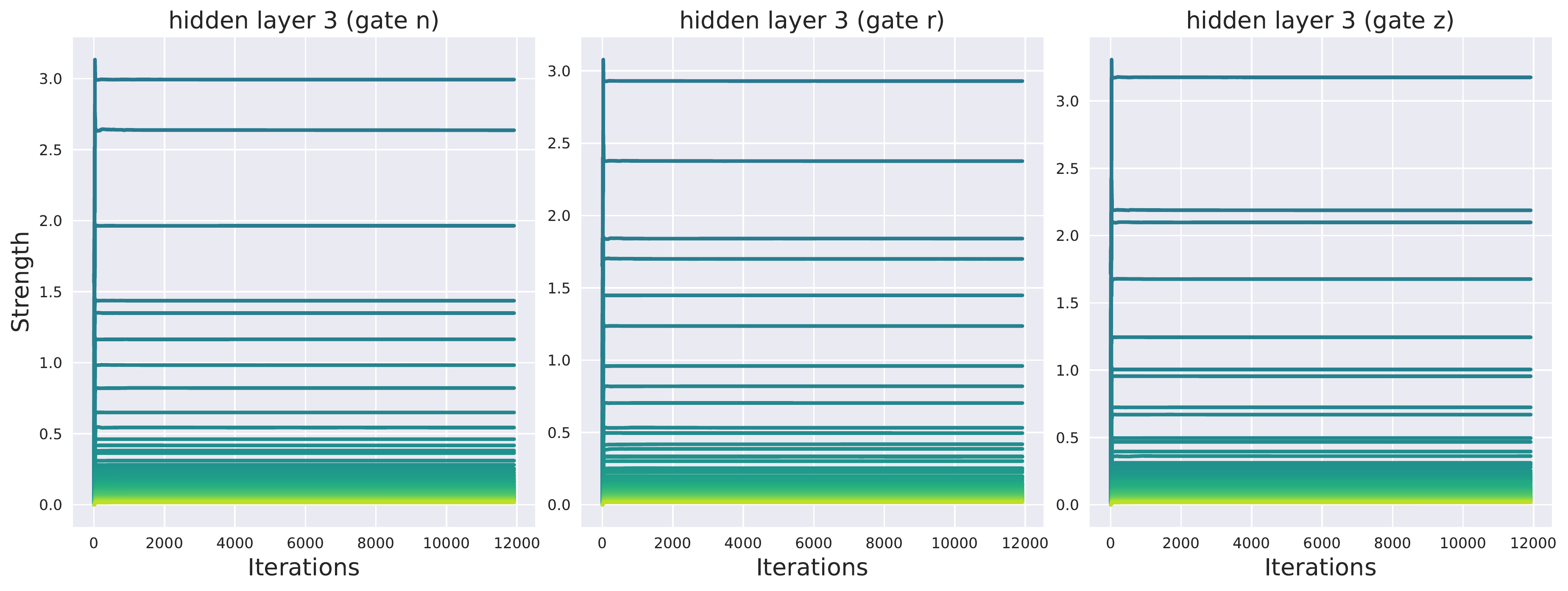}
\end{figure}
\subsubsection{Six depth LSTM}
\begin{figure}[H]
    \centering
    \includegraphics[width=1.0\linewidth]{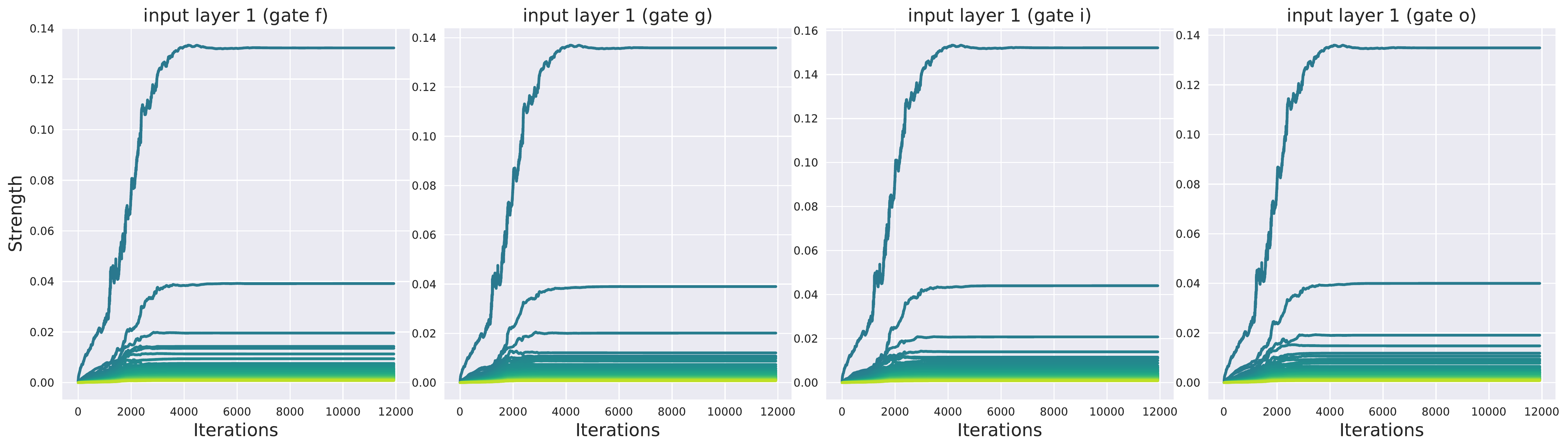}
    \includegraphics[width=1.0\linewidth]{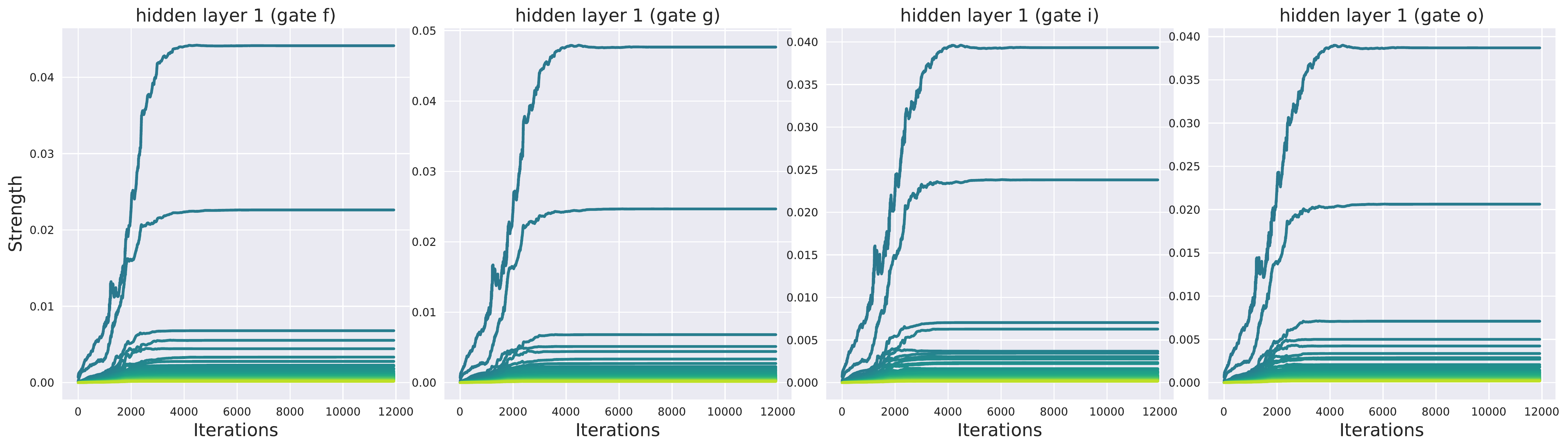}
    \includegraphics[width=1.0\linewidth]{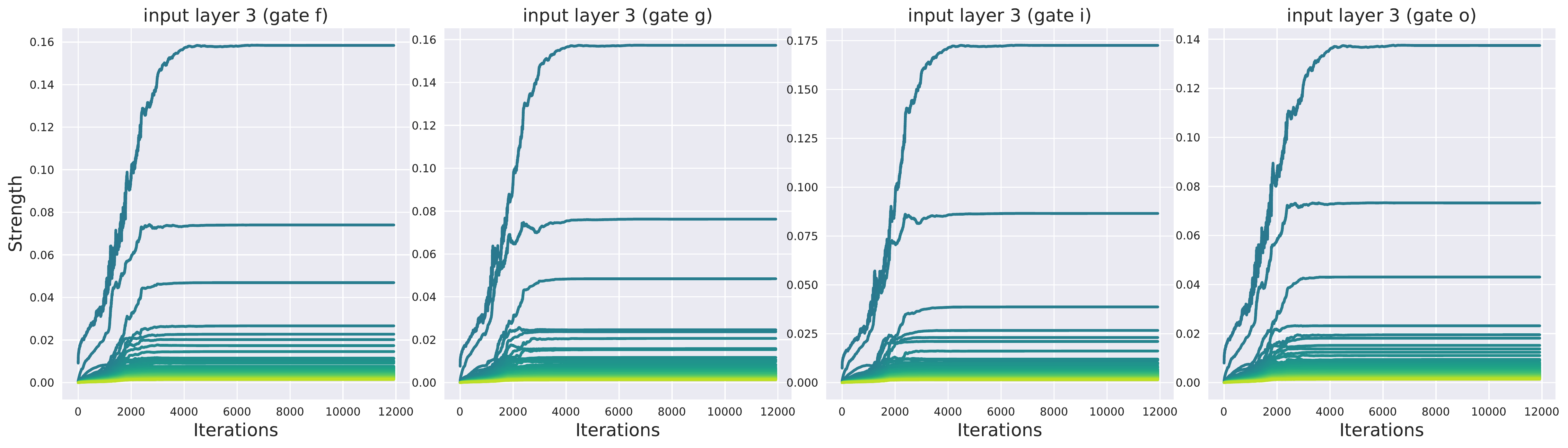}
    \includegraphics[width=1.0\linewidth]{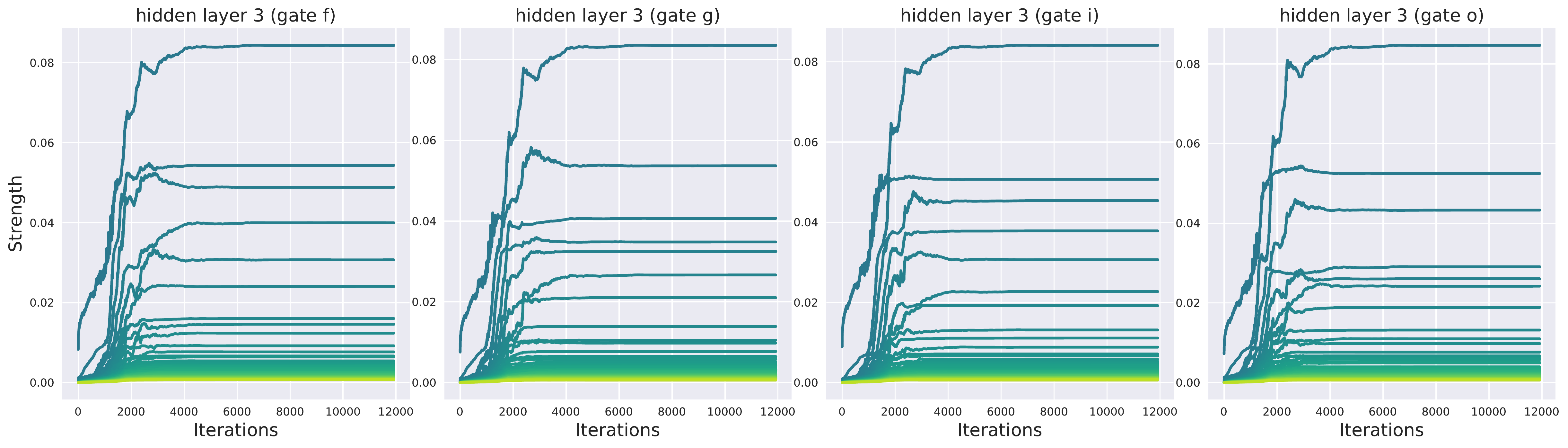}
\end{figure}
\subsection{Connection with generalization loss}
\begin{figure}[H]
    \centering
    \includegraphics[width=0.75\linewidth]{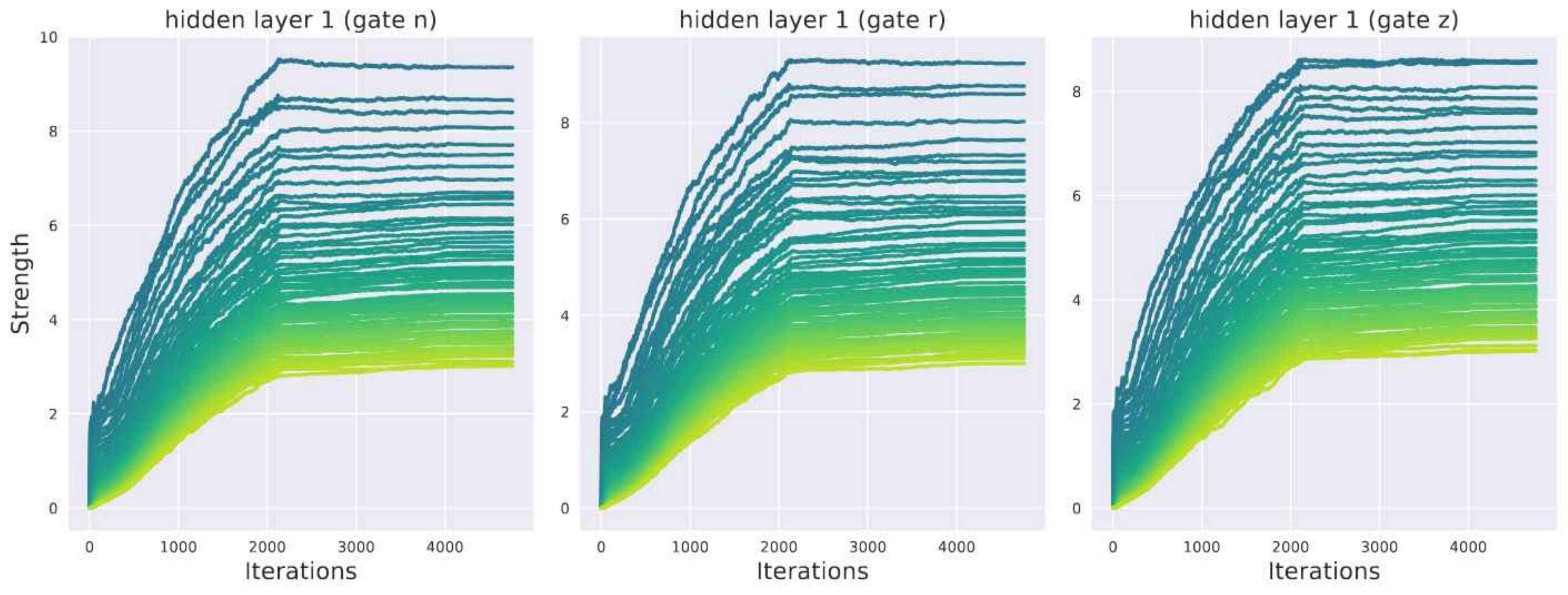}
    \includegraphics[width=0.75\linewidth]{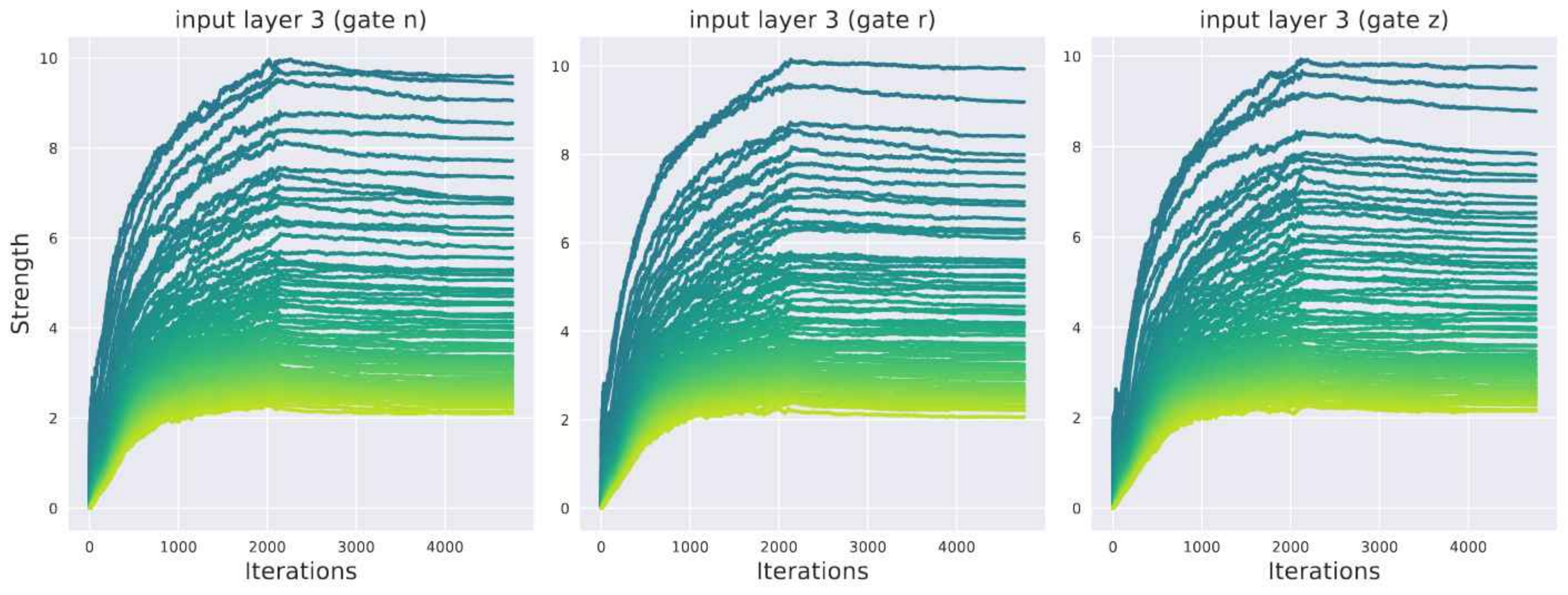}
    \includegraphics[width=0.75\linewidth]{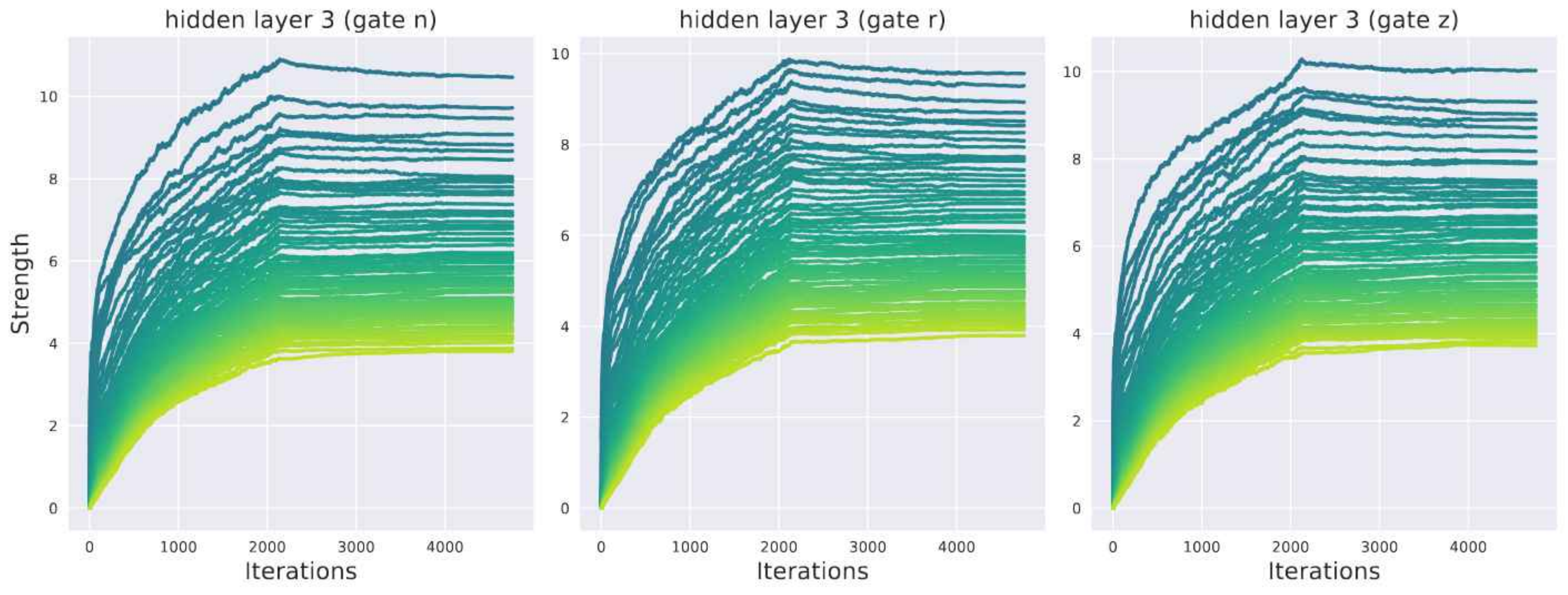}
    \includegraphics[width=0.25\linewidth]{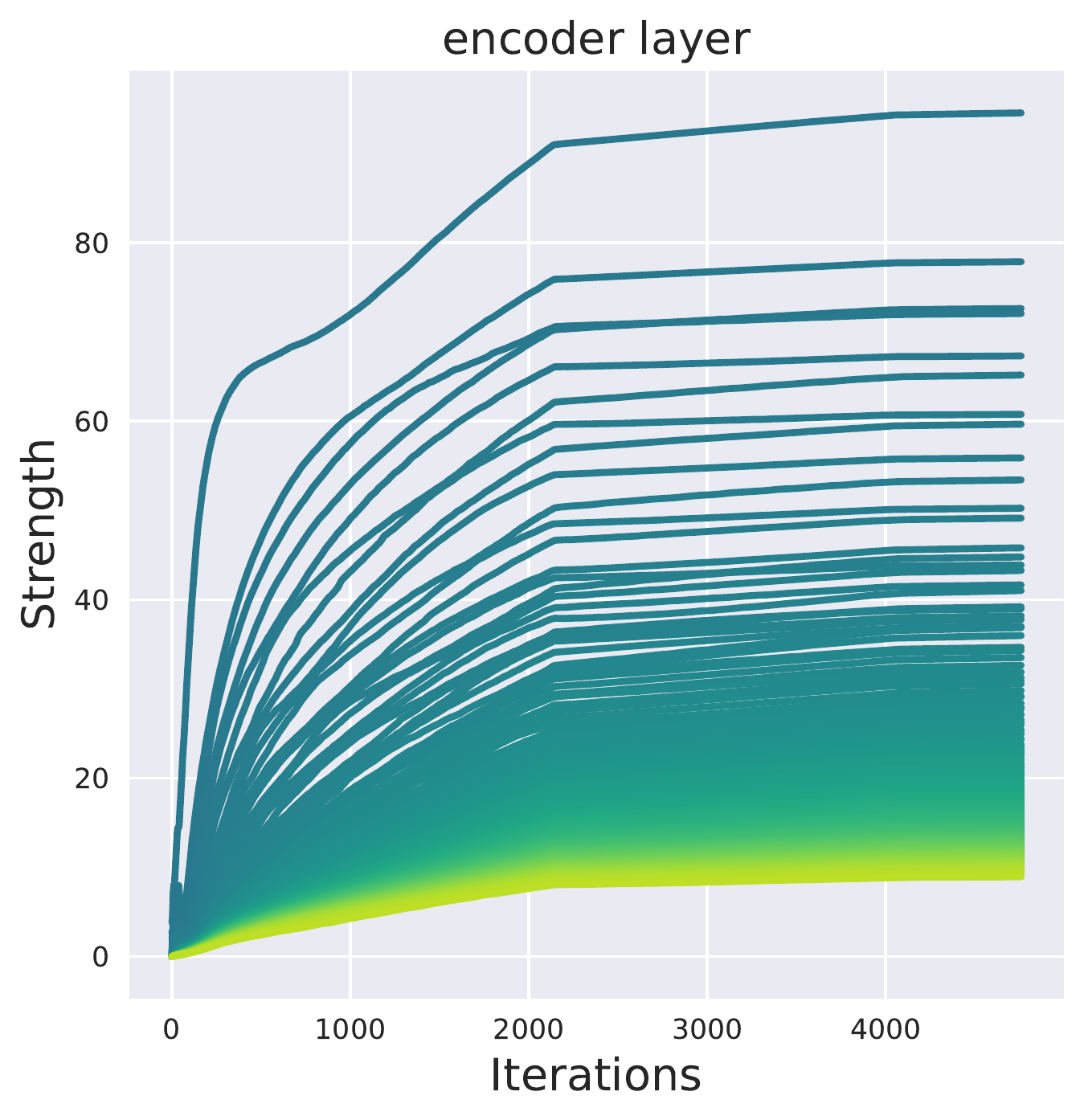}
    \caption{The evolutions of all singular vectors of cumulative weight updates $D_t$ over the training of 4-layer GRU after decreasing learning rate or enlarging batch size. To save space, we select layer 1 and layer 3 to display.}
    \label{fig:spectrum_over_training_GRU-4}
\end{figure}

However, we find being deeper makes generalization loss poor compared with shallow ones in GRU and LSTM. When we tune 4-layer GRU back to the normal generalization loss by decreasing learning rate or enlarging batch size, as shown in Figure~\ref{fig:spectrum_over_training_GRU-4}, the low-rank bias behaves not clearly again. This implies the interplay between the low-rank bias and the generalization. We hope these empirical results provide more insights into the complexity of neural network architecture and the design of low-rank algorithm in the cumulative weight updates, including LoRA \citep{hu_lora_2021}. 
\begin{figure}[H]
    \centering
    \includegraphics[width=0.75\linewidth]{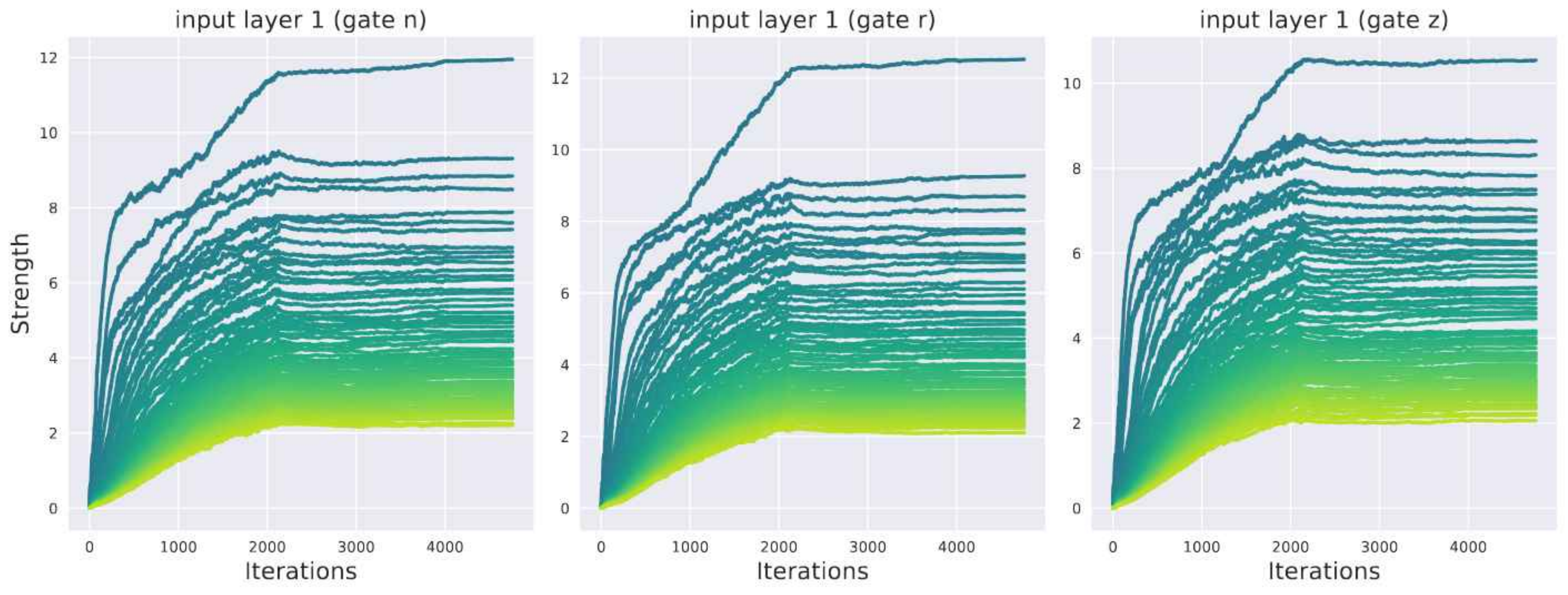}
\end{figure}

\subsection{Low-Rank learning under different learning algorithms}

\begin{figure}[H]
    \centering
    \includegraphics[width=0.75\linewidth]{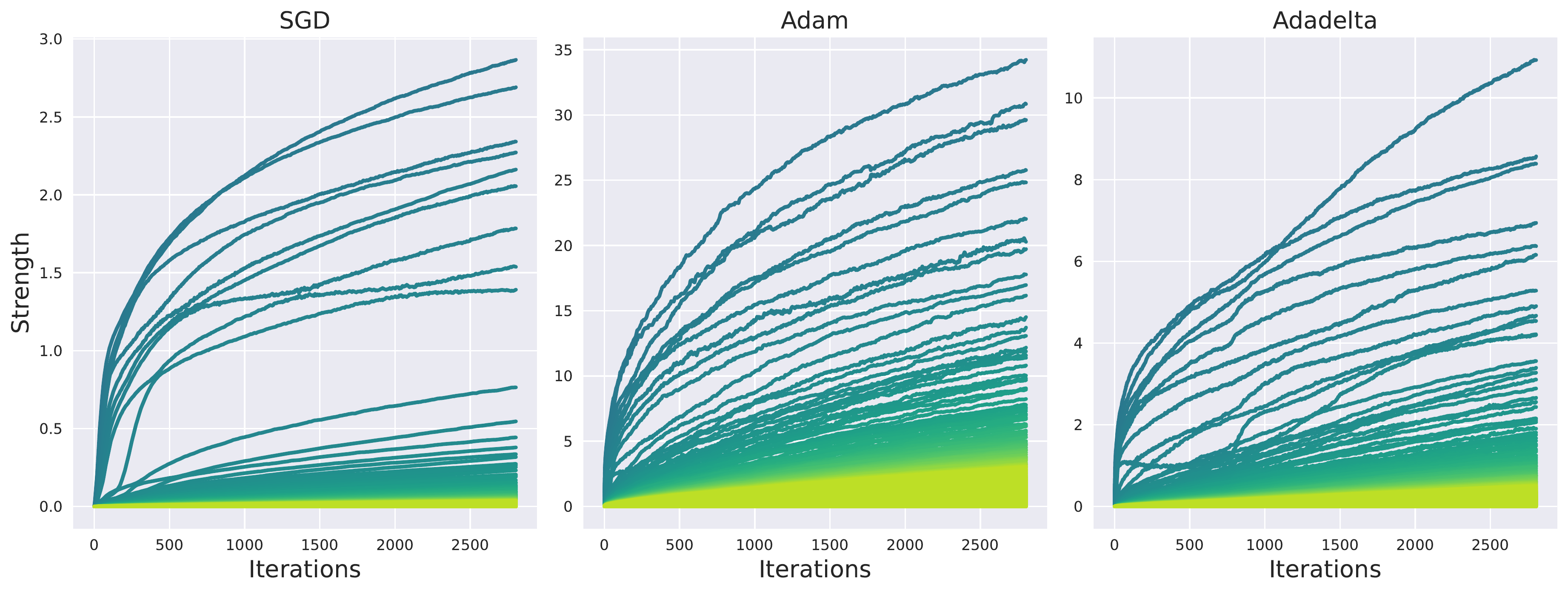}
    \caption{The evolutions of all singular vectors of cumulative weight updates $D_t$ over the training of MLP using SGD, Adam, and Adadelta optimizers. Darker colors indicate singular vectors with higher strengths.}
\end{figure}

\end{document}